\documentclass{article}

\PassOptionsToPackage{numbers, compress}{natbib}

\usepackage[preprint]{neurips_2026}

\usepackage[utf8]{inputenc} 
\usepackage[T1]{fontenc}    
\ifdefined\XeTeXversion
  \usepackage{newunicodechar}
  \newunicodechar{—}{\textemdash}
  \newunicodechar{‑}{\mbox{-}}
\fi
\usepackage{hyperref}       
\usepackage{url}            
\usepackage{booktabs}       
\usepackage{amsfonts}       
\usepackage{nicefrac}       
\usepackage{microtype}      
\usepackage{multirow}
\usepackage{graphicx}
\usepackage{amsmath}
\usepackage{amsthm}
\usepackage{bm}
\usepackage[table,xcdraw]{xcolor}
\definecolor{mygray}{gray}{0.85}
\newcommand{\gc}{\cellcolor{mygray}}

\theoremstyle{plain}
\newtheorem{theorem}{Theorem}[section]

\newtheorem{lemma}[theorem]{Lemma}
\newtheorem{corollary}[theorem]{Corollary}
\theoremstyle{definition}
\newtheorem{definition}[theorem]{Definition}

\theoremstyle{remark}

\usepackage{algorithm}
\usepackage{algpseudocode}
\usepackage{amssymb}
\usepackage{wrapfig}
\usepackage[capitalize]{cleveref}
\usepackage{float}
\crefname{figure}{Figure}{Figures}
\crefname{equation}{Equation}{Equations}

\title{ROSE: Risk-Aware Orthogonal Subspace Navigation for Lifelong Knowledge Editing in Multimodal Large Language Models}

\author{%
  \textbf{Lingyun Song$^{1,2}$\thanks{Corresponding authors.},
  Ziyao Chen$^{1}$,
  Kang Pan$^{3}$,
  Xiaolin Han$^{1,*}$}
  \\[0.2em]
  \textbf{Yifei Zhang$^{1}$,
  Xiaoqi Wang$^{1}$,
  Yudai Pan$^{1,*}$,
  Xuequn Shang$^{1,4,*}$}
  \\[0.7em]
  $^{1}$School of Computer Science, Northwestern Polytechnical University, Xi'an\\
  $^{2}$Shenzhen Research Institute of Northwestern Polytechnical University, Shenzhen\\
  $^{3}$Independent Researcher\\
  $^{4}$Key Laboratory of Big Data Storage and Management, Northwestern Polytechnical University,\\
  Ministry of Industry and Information Technology, Xi'an\\
  \texttt{lysong@nwpu.edu.cn, xiaolinh@nwpu.edu.cn}\\
  \texttt{panyudai@nwpu.edu.cn, shang@nwpu.edu.cn}
}
\begin{document}

\maketitle

\begin{abstract}

Multimodal Large Language Models (MLLMs) require continuous factual updates to stay current, yet lifelong knowledge editing remains a daunting challenge. The primary obstacle is the severe Locality Erosion and Catastrophic Forgetting, where sequential edits inevitably interfere with pre-trained capabilities and prior updates. In this paper, we propose \textbf{ROSE}, a \textbf{R}isk-aware \textbf{O}rthogonal \textbf{S}ubspace navigation framework for lifelong knowledge \textbf{E}diting. ROSE addresses forgetting by aiming to restrict parameter updates to the orthogonal complement of previously edited subspaces. Crucially, it introduces a risk-guided mechanism that utilizes a dynamically computed risk mask to safeguard specific factual parameter subspaces essential for model stability. During inference, ROSE employs a multi-granularity knowledge integration strategy, featuring prototype-anchored semantic gating and perturbation-driven subspace fusion to precisely activate edited knowledge for relevant queries while limiting interference on unrelated inputs. Comprehensive experiments on two prominent multimodal benchmarks show that ROSE is competitive with state-of-the-art approaches across five key metrics, while maintaining outstanding stability over $1,000$ sequential updates. Our code is available at \url{https://github.com/sunlitsong/ROSE}.
\end{abstract}

\section{Introduction}

Multimodal Large Language Models (MLLMs) have demonstrated remarkable capabilities in advanced reasoning and cross-modal understanding, largely due to their ability to align visual and textual semantics \citep{9827595} and fuse information from diverse data sources \citep{comanici2025gemini, yang2025qwen3, achiam2023gpt, liu2023visual}. However, these models remain tethered to the static knowledge captured during their pre-training on fixed datasets. This ``knowledge freezing'' limits their ability to adapt to a rapidly evolving world, frequently resulting in hallucinated content and outdated responses when faced with new events \citep{wang2024knowledge, zhang2024comprehensive}. While conventional full-parameter fine-tuning is a natural consideration for updates, it is computationally prohibitive and prone to catastrophic forgetting \citep{zheng2025towards, zhai2023investigating}. Furthermore, most existing model editing methods are designed for static, isolated updates \citep{meng2022locating, mitchell2021fast, meng2022mass}, making them ill-suited for the continuous, sequential data streams encountered in real-world scenarios.

\begin{wrapfigure}{!t}{0.42\textwidth}
    \centering
    \includegraphics[width=0.40\textwidth]{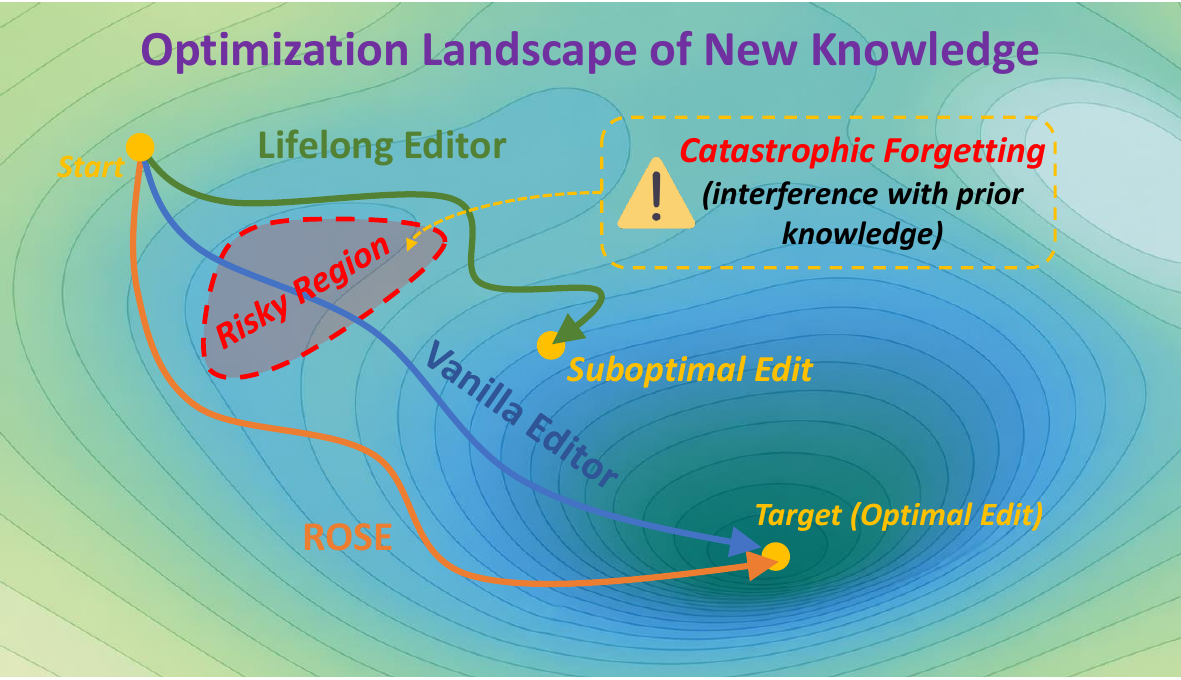} 
    \caption{Comparison of optimization trajectories generated by conventional Editors and our ROSE. By navigating parameter updates away from high-risk general subspaces, ROSE achieves a superior balance between incorporating new knowledge and preserving pre-trained stability.}
    \label{fig:compare}
\end{wrapfigure}

Consequently, the capacity for Lifelong Knowledge Editing (LKE) has become essential, requiring not only efficient modification of specific facts but also long-term reliability across a prolonged stream of knowledge \citep{chen2024lifelong, wang2024wise, hartvigsen2023aging}. In the multimodal domain, this task faces two challenges: (1) \textbf{Catastrophic Forgetting}, where new updates disrupt previously learned edits, and (2) \textbf{Locality Erosion}, where the model's performance on unrelated knowledge deteriorates. These issues are exacerbated by the complex modal entanglement between visual and textual representations. Such entanglement makes the model's knowledge space particularly sensitive to sequential perturbations, where a single factual update can inadvertently trigger a chain reaction of degradation across different modalities \citep{cheng2023can, zhang2024comprehensive}.

The fundamental issue lies in the unresolved tension between editing effectiveness and knowledge preservation. As illustrated in \cref{fig:compare}, existing lifelong editors struggle to navigate this trade-off: they are often either overly conservative, which reduces the reliability of new edits, or overly intrusive, leading to knowledge interference that corrupts both pre-trained capabilities and prior edits \citep{wang2024lemoe, chen2025lifelong}. Crucially, these approaches lack subspace awareness, as they provide no guidance for pinpointing which parameter subspaces are susceptible to degradation. Consequently, as edits accumulate, the boundaries of modified knowledge become increasingly blurred, leading to ``false triggers'' where edited knowledge is unintentionally activated for irrelevant inputs, thereby undermining model locality \citep{mitchell2022memory, chen2024lifelong}.

In this paper, we introduce ROSE, a new framework for Risk-aware Orthogonal Subspace navigation designed for lifelong knowledge Editing in MLLMs. ROSE operates on a ``navigate-then-integrate'' philosophy to safeguard the model's foundational capabilities during continuous updates. During the editing phase, it navigates weight updates into the orthogonal complement of previously stored knowledge, while employing a risk mask to anchor parameter subspaces essential for general stability. During the inference phase, it integrates the corresponding knowledge through a multi-granularity knowledge integration strategy, using prototype similarity \citep{11397820} to assess input relevance. By combining prototype-anchored gating with perturbation-driven fusion, ROSE selectively activates edited knowledge for pertinent inputs while routing queries rejected by the gate to the stable pre-trained backbone. This dual-phase approach ensures both high plasticity for new facts and robust reliability for existing capabilities. Our key contributions can be outlined as follows:

\begingroup
\setlength{\topsep}{4pt}
\setlength{\partopsep}{0pt}
\begin{itemize}
    \setlength{\itemsep}{5pt}
    \setlength{\parsep}{0pt}
    \setlength{\parskip}{0pt}
    \item \textbf{Risk-Aware Orthogonal Editing Framework}: We introduce ROSE, a novel framework for lifelong MLLM editing. It features a risk-guided orthogonal editing mechanism that aims to restrict parameter updates to the orthogonal complement of the existing knowledge subspace. This is augmented by a dynamic risk mask that anchors parameters critical for general pre-trained capabilities, effectively mitigating catastrophic forgetting.
    
    \item \textbf{Multi-Granularity Knowledge Integration}: We propose a coarse-to-fine integration strategy to bridge the gap between pre-trained and edited knowledge. By synergizing prototype-anchored semantic gating with perturbation-driven subspace fusion, this module adaptively routes and integrates representations during inference, significantly mitigating locality erosion by preventing unintended factual triggers.
    
    \item \textbf{Extensive Empirical Validation}: We conduct extensive evaluations on two prominent multimodal benchmarks. The results across five key metrics demonstrate that ROSE achieves state-of-the-art performance.%
    
\end{itemize}
\endgroup

\section{Background and Related works}
\label{sec:related_works}

\subsection{Multimodal Large Language Models}

Multimodal Large Language Models (MLLMs) have achieved remarkable strides in visual perception and linguistic understanding, including knowledge-based visual question answering \citep{SONG2023109015} and image captioning \citep{yang2025qwen3, liu2023visual, achiam2023gpt, li2023blip}. To integrate complementary information across modalities \citep{ZHU2026103981}, these models typically adopt a modular architecture comprising a vision encoder \citep{radford2021learning, dosovitskiy2020image}, a vision-to-language projection module \citep{li2023blip, lin2022cat}, and a Large Language Model (LLM) backbone \citep{touvron2023llama, chung2024scaling}. Despite their prowess, MLLMs rely on static training datasets, causing their internal knowledge to become rapidly outdated. This limitation often manifests as hallucinations or incorrect responses to newly emerging facts. Crucially, the complex interplay between modalities \citep{SONG2025103082} suggests that factual knowledge is not stored in isolation, but rather resides within specific parametric subspaces, a characteristic that motivates our subspace-based navigation approach.

\subsection{Knowledge Editing}

Knowledge editing seeks to modify specific factual associations within a model efficiently without retraining from scratch. Current paradigms can be categorized into four groups: (1) Locate-and-Edit techniques, such as ROME \citep{meng2022locating} and MEMIT \citep{meng2022mass}, which pinpoint and update neurons encoding specific facts; (2) Meta-Learning approaches like MEND \citep{mitchell2021fast}, which employ hypernetworks to predict weight updates from edit gradients; (3) Parameter-Efficient methods \citep{hartvigsen2023aging, dong2022calibrating} that attach trainable adapters to a frozen backbone; and (4) Retrieval-based methods \citep{mitchell2022memory, zheng2023can} that steer model behavior via in-context learning and forgo any parameter changes. While effective for isolated updates, these paradigms often lack a strategy to mitigate the overlapping interference that accumulates across a sequence of edits. This leads to a progressive degradation of both new and old knowledge, a limitation that necessitates a more robust mechanism for parameter isolation in lifelong scenarios.

\subsection{Lifelong Knowledge Editing}

Lifelong knowledge editing seeks to continually revise models while alleviating sequential catastrophic forgetting. Existing research primarily explores parameter isolation and modular expansion. For instance, RECIPE \citep{chen2024lifelong} utilizes sentinel-retrieved prompts to disentangle edits, while LEMoE \citep{wang2024lemoe} employs Mixture-of-Experts (MoE) architectures to distribute knowledge across different experts. Although these strategies have shown promise, scaling them to MLLMs remains difficult due to complex cross-modal entanglement, which LiveEdit \citep{chen2025lifelong} attempts to address by employing distinct routing mechanisms. However, while routing facilitates knowledge localization, these modular approaches often overlook the internal optimization trajectory, failing to prevent subspace collisions during the update of shared parameters or protect factual subspaces critical for pre-trained knowledge. ROSE fills this gap by introducing risk-aware orthogonal navigation, providing a rigorous geometric and risk-based constraint to safeguard model integrity at the parameter level.

\section{Preliminaries}

In this section, we provide a formal definition of the multimodal lifelong knowledge editing task and its associated evaluation metrics, following the conventions established in \cite{huang2024vlkeb} and \cite{zheng2023can}.

\subsection{Multimodal Lifelong Knowledge Editing}

Let $f(v, p; \theta)$ denote a MLLM parameterized by $\theta$, where $v \in \mathcal{V}$ is an input image and $p \in \mathcal{P}$ is a textual prompt. The model generates an output $o = f(v, p; \theta)$. 

\textbf{Knowledge Editing}. The goal of knowledge editing is to modify the model's behavior on a specific factual instance $e = (v_e, p_e, o_e \to o^*_e)$, where $o_e$ is the original (incorrect or outdated) response and $o^*_e$ is the target correct output. An editor $\mathcal{E}(f, e)$ aims to produce an updated parameter set $\theta'$ such that $f(v_e, p_e; \theta') = o^*_e$.

\textbf{Lifelong Setting}. In the lifelong knowledge editing scenario, the model encounters a continuous stream of edits $\mathcal{D}_{\text{edit}} = \{e_1, e_2, \dots, e_T\}$. At each time step $t$, the model updates its parameters from $\theta_{t-1}$ to $\theta_t$ using the current edit $e_t$. The fundamental difficulty lies in managing the sequential interference within the parameter space, where updates for $e_t$ may catastrophically disrupt the representations learned for prior edits or erode the model's inherent pre-trained capabilities.

\subsection{Evaluation Metrics}
We evaluate ROSE across three core dimensions: (1) \textbf{Reliability} quantifies how effectively the model incorporates the specific target knowledge it was modified to handle; (2) \textbf{Generality} evaluates whether the edited knowledge generalizes to semantically equivalent samples within a neighborhood, assessed via rephrased texts ({T-Gen.}) and visually transformed images ({M-Gen.}); and (3) \textbf{Locality} safeguards the integrity of the model's original pre-trained knowledge against unintended interference on unrelated textual ({T-Loc.}) and multimodal ({M-Loc.}) samples. The formal mathematical definitions for these metrics are provided in the Appendix \ref{app:metrics}.

\begin{figure}[]
    \centering
    \includegraphics[width=1\linewidth]{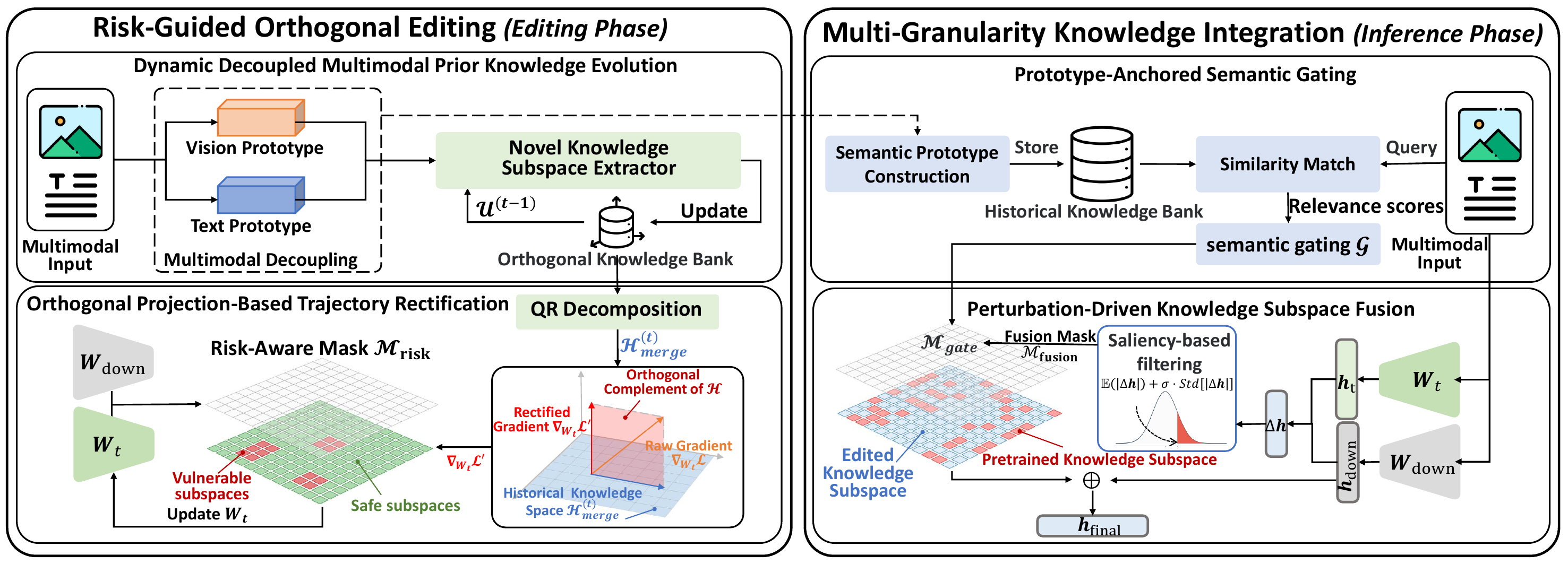} 
    \caption{An illustration of the ROSE framework. It mainly consists of two core modules: the Risk-Guided Orthogonal Editing and the Multi-Granularity Knowledge Integration.}
    \label{fig:model}
\end{figure}

\section{Methodology}

In this section, we introduce the ROSE framework, designed to alleviate two distinct types of catastrophic forgetting in lifelong editing: the degradation of previously applied edits and the erosion of the model's inherent pre-trained capabilities. As illustrated in Fig.\ref{fig:model}, ROSE consists of two synergistic modules: (1) a Risk-Guided Orthogonal Editing mechanism for safe knowledge absorption, and (2) a Multi-Granularity Knowledge Integration strategy for adaptive knowledge retrieval during inference. The complete procedural workflow is summarized in Appendix \ref{sec:algorithm_details}.

\subsection{Risk-Guided Orthogonal Editing Mechanism}\label{sec:module1}

Factual knowledge primarily resides in the MLP down-projection matrices. We therefore target $\boldsymbol{W}_{\text{down}} \in \mathbb{R}^{d_{m} \times d_{n}}$ at the $l$-th layer, where $d_m$ and $d_n$ represent the output and input dimensions, respectively. To preserve pre-trained capabilities, we freeze the backbone and isolate sequential updates within an auxiliary branch. Let $\boldsymbol{W}_t$ denote the parameters of this auxiliary branch at editing step $t$, initialized as $\boldsymbol{W}_0 = \boldsymbol{W}_{\text{down}}$ before any edits occur.

For the $t$-th edit $e_t=(v_t,p_t,o_t)\in\mathcal{D}_{\text{edit}}$, let $\boldsymbol{x}^{(t)} \in \mathbb{R}^{d_n \times L}$ denote its input features, where $L$ denotes the token length. To prevent catastrophic forgetting of prior edits, the ideal weight update $\Delta \boldsymbol{W}_t$ aims to satisfy the following orthogonal constraint:
\begin{equation}
    (\boldsymbol{W}_t+\Delta \boldsymbol{W}_t) \boldsymbol{x}_{\text{pre}} = \boldsymbol{h}_{\text{pre}},
    \label{eq:constraint}
\end{equation}
where $\boldsymbol{x}_{\text{pre}}$ and $\boldsymbol{h}_{\text{pre}}$ represent the input and output features of the edited samples, respectively. Given that $\boldsymbol{W}_t \cdot \boldsymbol{x}_{\text{pre}} = \boldsymbol{h}_{\text{pre}}$ already holds, this reduces to requiring $\Delta \boldsymbol{W}_t$ to lie in the null space of past inputs, i.e., $\Delta \boldsymbol{W}_t \cdot \boldsymbol{x}_{\text{pre}} = \boldsymbol{0}$.

\textbf{Dynamic Decoupled Multimodal Prior Knowledge Evolution.} 
To alleviate cross-modal entanglement, we decouple $\boldsymbol{x}^{(<t)}$ into a visual subspace $\mathcal{H}_v^{(<t)}$ and a textual subspace $\mathcal{H}_p^{(<t)}$. Specifically, we decompose each past input $\boldsymbol{x}^{(t-1)}$ into visual tokens $\boldsymbol{x}^{(t-1)}_v$ and textual query tokens $\boldsymbol{x}^{(t-1)}_p$. Then we encode the visual modality by the centroid $\boldsymbol{x}'_v = \frac{1}{L_v}\sum_{i=1}^{L_v} \boldsymbol{v}_i$ and the textual modality by the final query token $\boldsymbol{x}'_p$. This aggregation balances semantic expressiveness and computational efficiency.

To effectively manage the lifelong accumulation of knowledge over time, we maintain an orthogonal matrix $\boldsymbol{U_*} \in \mathbb{R}^{d_n \times t}$ that spans the existing knowledge subspace $\mathcal{H}_*$. Here, $*\in\{v,p\}$. When a new edit feature $\boldsymbol{x}_*' \in \{\boldsymbol{x}'_p, \boldsymbol{x}'_v\}$ arrives, we first normalize it as $\hat{\boldsymbol{x}}_* = \boldsymbol{x}_*' / \|\boldsymbol{x}_*'\|_2$. To embed this new direction into the existing subspace while maintaining orthogonality, we project $\hat{\boldsymbol{x}}_*$ onto the orthogonal complement of $\boldsymbol{U_*}^{(t-1)}$, yielding a residual $\tilde{\boldsymbol{x}}_*$ that is independent of all preceding edits:
\begin{equation}
\tilde{\boldsymbol{x}}_* = \hat{\boldsymbol{x}}_* - \boldsymbol{U_*}^{(t-1)} (\boldsymbol{U_*}^{(t-1)})^\top \hat{\boldsymbol{x}}_*.
\end{equation}
This process effectively filters out the information already captured by earlier edits, so that $\boldsymbol{U_*}$ becomes a matrix whose columns are the existing orthonormal basis vectors. To ensure numerical stability and keep the subspace compact, we only expand the basis when the $L_2$ norm of the residual is greater than a predefined threshold $\epsilon$:
\begin{equation}
\boldsymbol{U_*}^{(t)} \leftarrow 
\begin{cases} 
\left[ \boldsymbol{U_*}^{(t-1)}, \frac{\tilde{\boldsymbol{x}}_*}{\Vert\tilde{\boldsymbol{x}}_*\Vert_2} \right], & \text{if } \Vert\tilde{\boldsymbol{x}}_*\Vert_2 > \epsilon. \\
\boldsymbol{U_*}^{(t-1)}, & \text{otherwise}.
\end{cases}
\end{equation}

\textbf{Orthogonal Projection-Based Trajectory Rectification.} By iteratively updating $\boldsymbol{U_*}$ through this subspace-aware mechanism, we ensure it effectively captures the edited core knowledge space. To satisfy the constraint $\Delta \boldsymbol{W}_t \cdot \boldsymbol{x}_{\text{pre}} = \boldsymbol{0}$ established in \cref{eq:constraint}, the weight update must be restricted to the null space of $\mathcal{H}_*$. Because $\boldsymbol{U}_v$ and $\boldsymbol{U}_p$ arise from distinct modalities, we combine them into a single representation $\boldsymbol{U}_{\text{merged}}^{(t)} = [\boldsymbol{U}_{v}^{(t)}, \boldsymbol{U}_{p}^{(t)}]$. We then apply a QR decomposition to this merged matrix to obtain a strictly orthonormal basis for the resulting subspace:
\begin{equation}
\boldsymbol{Q}^{(t)}, \boldsymbol{R}^{(t)} = \text{QR}(\boldsymbol{U}_{\text{merged}}^{(t)}),
\end{equation}
where $\boldsymbol{Q}^{(t)} \in \mathbb{R}^{d_n \times 2t}$ denotes an orthonormal basis of the prior knowledge subspace $\mathcal{H}_v^{(t)} \bigcup \mathcal{H}_p^{(t)}$. 

While projecting gradients away from $\boldsymbol{Q}^{(t)}$ helps protect explicit prior edits, it leaves the vast pre-training distribution vulnerable. Inspired by Task Vectors \cite{ilharco2022editing}, we quantify parameter vulnerability via absolute weight subtraction, as large deviations naturally highlight the specific parameters responsible for acquiring new facts. To preserve inherent general capabilities, we restrict updates to only the most active parameters. Compared to complex gradient-based metrics, this highly efficient strategy effectively pinpoints the factual parameter subspace using a risk mask $\boldsymbol{\mathcal{M}}_{\text{risk}}$:
\begin{equation}
\label{equ:mask-pert}
\boldsymbol{\mathcal{M}}_{\text{risk}} = \Bigl[ \mathbb{I}\bigl( |\Delta \boldsymbol{W}_{i,j}| > \psi_k \bigr) \Bigr]_{i,j},
\end{equation}
where $\Delta \boldsymbol{W} = \boldsymbol{W}_{t} - \boldsymbol{W}_{\text{down}}$, $\psi_k$ is the $(1-k)$-th quantile of absolute differences $\{|\Delta \boldsymbol{W}_{i,j}|\}$, and $k \in (0,1]$ dictates the selection fraction. Using the value captures weight plasticity, and employing $\boldsymbol{W}_{\text{down}}$ as a static anchor prevents cumulative error drift (further discussed in Appendix \ref{app:more_exp}).

Next, we integrate the orthogonal constraint with the empirically motivated risk mask to rectify the raw optimization gradient $\nabla_{\boldsymbol{W}_t} \mathcal{L}$:
\begin{equation}
\label{eq:grad_update}
\nabla_{\boldsymbol{W}_t} \mathcal{L} \leftarrow \left(\nabla_{\boldsymbol{W}_t} \mathcal{L} - (\nabla_{\boldsymbol{W}_t} \mathcal{L} \cdot\boldsymbol{Q}^{(t)}) (\boldsymbol{Q}^{(t)})^\top\right)\odot\boldsymbol{\mathcal{M}}_{\text{risk}},
\end{equation}
where $\mathcal{L}$ is the loss function, and $\odot$ is the Hadamard product. Crucially, $\boldsymbol{\mathcal{M}}_{\text{risk}}$ is recomputed at each gradient step from the current $\boldsymbol{W}_t$. The inner projection steers gradients away from the prior knowledge subspace $\boldsymbol{Q}^{(t)}$, while $\boldsymbol{\mathcal{M}}_{\text{risk}}$ structurally regularizes the update by suppressing parameters critical for pre-trained capabilities.

\subsection{Multi-Granularity Knowledge Integration Strategy}\label{sec:module2}
To mitigate erosion of pretrained knowledge, we propose a multi-granularity routing mechanism that dynamically determines the scope of each edit and integrates relevant knowledge spaces during inference, thereby maximizing the preservation of the model's locality.

\textbf{Semantic Prototype Construction.} To maintain a persistent record of the semantic scope for each update, we establish a Knowledge Bank $\mathcal{K}_*$. For each edit $e$, we derive its prototype $\boldsymbol{k}_*$ by aggregating the representations of the corresponding modality tokens. Specifically, for an input $\boldsymbol{x} \in \mathbb{R}^{d_n \times L}$ during the editing phase, we update this bank by concatenating the normalized centroid:
\begin{equation}
    \boldsymbol{k}_* = \frac{\sum_{i \in \mathcal{I}_*} \boldsymbol{x}_i}{\left\| \sum_{i \in \mathcal{I}_*} \boldsymbol{x}_i \right\|_2}, \quad \mathcal{\boldsymbol{K}}_* \leftarrow [\mathcal{\boldsymbol{K}}_*, \boldsymbol{k}_*],
\end{equation}
where $\mathcal{I}_*$ denotes the set of token indices belonging to modality $* \in \{p, v\}$ and $\Vert\cdot\Vert_2$ denotes the $L_2$ normalization operator. This cumulative structure allows the model to preserve the individual semantic representation of each edit without the information compression inherent in orthogonalization.

At inference time, for any given input $\boldsymbol{x}$, we begin by assessing how relevant it is to the edited knowledge. To do this, we derive a global query vector $\boldsymbol{q}_*$ by computing the centroid specific to its modality: $\boldsymbol{q}_* = \sum_{i \in \mathcal{I}_*} \boldsymbol{x}_i/\left\Vert \sum_{i \in \mathcal{I}_*} \boldsymbol{x}_i\right\Vert_2$.

\textbf{Prototype-Anchored Semantic Gating.} Acting as a semantic router, this module directs inputs by quantifying their semantic affinity to prior knowledge subspaces $\mathcal{H}_p$ and $\mathcal{H}_v$. We compute the maximal projection alignment of the $L_2$-normalized query $\boldsymbol{q}_*$ onto the directional bases in the knowledge bank $\boldsymbol{\mathcal{K}}_*$. Formally, the relevance score $s_*$ is defined as the maximum cosine similarity over the stored prototypes:
\begin{equation}
    s_* = \max_{j \in \{1, \dots, N\}} \langle \boldsymbol{q}_*, \boldsymbol{k}_{*,j} \rangle,  \quad \text{where } * \in \{v, p\}.
\end{equation}
where $\langle \cdot, \cdot \rangle$ denotes the inner product, and $N$ denotes the current number of prototypes stored in $\mathcal{K}_*$. It essentially captures the highest possible alignment between the input features and any single basis vector in the knowledge subspace, and guarantees that the router activates the auxiliary branch only when the input lies sufficiently close to a previously modified knowledge subspace.

The semantic gating $\mathcal{G}\in \{0, 1\}$ is then formulated by
\begin{equation}
    \mathcal{G} = \mathbb{I}\Big( 
        \bigwedge_{*\in\{v,p\}} \bigl(s_* > \tau_{\min}\bigr) 
        \;\land\; 
        \Big( \bigvee_{*\in\{v,p\}} \bigl(s_* > \tau\bigr)
        \;\lor\; |s_v - s_p| < \tau_{\text{gap}}
    \Big)    \Big),
\end{equation}
where $\tau_{\text{min}}$ and $\tau_{\text{gap}}$ serve as safeguards that exclude only obviously irrelevant or cross-modally inconsistent inputs, leaving $\tau$ as the principal semantic threshold that governs the routing behavior. With this mechanism, the auxiliary branch is activated only when the input demonstrates strong linguistic or visual correspondence with the edited distribution.

\textbf{Perturbation-Driven Knowledge Subspace Fusion.} Since factual knowledge in MLLMs tends to be encoded in distinct subspaces, we capture the output discrepancy between the two branches as $\Delta\boldsymbol{h} = \boldsymbol{h}_{t} - \boldsymbol{h}_{\text{down}} \in\mathbb{R}^{d_m \times L}$, where $\boldsymbol{h}_t$ and $\boldsymbol{h}_{\text{down}}$ are the outputs of $\boldsymbol{W}_t$ and $\boldsymbol{W}_{\text{down}}$ for the same input. Under the linear logit mapping analyzed in Appendix~\ref{app:proof_omission}, the first-order sensitivity analysis in Corollary~\ref{cor:kl_bound} motivates using per-dimension perturbation magnitudes $|\Delta\boldsymbol{h}_{i,j}|$ as a tractable criterion for selective fusion. Accordingly, we construct the perturbation-driven fusion mask $\boldsymbol{\mathcal{M}}_{\text{fusion}} \in \{0, 1\}^{d_m \times L}$ by retaining only those knowledge subspaces whose perturbation exceeds a data-driven threshold:
\begin{equation}
    \boldsymbol{\mathcal{M}}_{\text{fusion}} = \left[\mathbb{I} \left( |\Delta \boldsymbol{h}_{i,j}| > \mathbb{E}[|\Delta \boldsymbol{h}|] + \sigma \cdot \text{Std}[|\Delta \boldsymbol{h}|] \right)\right]_{i,j},
\end{equation}
where $\sigma$ is a hyperparameter, and $\mathbb{E}$ and $\text{Std}$ represent the expectation and standard deviation, respectively. The integrated output is regulated by a mask $\boldsymbol{\mathcal{M}}_{\text{gate}}=\mathcal{G} \cdot \boldsymbol{\mathcal{M}_{\text{fusion}}} \in \{0, 1\}^{d_m \times L}$, which in turn enables the integration of pretrained and edited knowledge:
\begin{equation}
\begin{aligned}
        \boldsymbol{h}_{\text{final}} &= \boldsymbol{h}_{\text{down}} + \boldsymbol{\mathcal{M}}_{\text{gate}} \odot \Delta \boldsymbol{h} \\
        &= (1-\boldsymbol{\mathcal{M}}_{\text{gate}})\odot\boldsymbol{h}_{\text{down}} + \boldsymbol{\mathcal{M}}_{\text{gate}} \odot \boldsymbol{h}_t,
\end{aligned}
\end{equation}
where $\odot$ denotes the Hadamard product. This hierarchical design allows the model to selectively modify particular knowledge subspaces for the edited facts, while preserving the original pre-trained representations across most of the overall knowledge space.

\section{Experiments}
\label{sec:exp}
\subsection{Experimental Setup}

\textbf{Benchmarks.} Following \cite{chen2025lifelong}, we assess ROSE on two widely used benchmarks: (1) MMEdit \cite{cheng2023can}, the first multimodal editing benchmark that includes both VQA and image captioning; and (2) VLKEB \cite{huang2024vlkeb}, a comprehensive knowledge editing benchmark constructed with real-world images for VQA, aimed at providing a challenging and high-quality evaluation. We report the image-captioning and visual question-answering tasks of MMEdit as E-IC and E-VQA, respectively. Details of the experimental setup are presented in Appendix \ref{app:exp_settings}.

\textbf{Baselines.} We select widely used and state-of-the-art approaches as baselines, spanning fine-tuning-based methods, meta-learning approaches \cite{mitchell2021fast}, retrieval-augmented methods \cite{mitchell2022memory}, lifelong editing frameworks \cite{chen2024lifelong, wang2024lemoe, chen2025lifelong}, as well as several other advanced techniques. Our experiments are conducted on a range of MLLM backbones, including LLaVA-v1.5 (7B, 13B) \cite{liu2023visual}, MiniGPT-4 \cite{zhu2023minigpt}, and Qwen2-VL \cite{wang2024qwen2}. Owing to space limitations, \cref{sec:main_results} mainly presents results on LLaVA-v1.5-7B, while the results for the remaining backbones are deferred to \cref{sec:backbone_scale}.

\begin{table*}[]
\caption{Performance comparison (\%) on LLaVA-v1.5 (7B), with baseline results taken from \citet{chen2025lifelong}. Models are assessed on five primary aspects: Reliability (Rel.), Text Generality (T-Gen.), Multimodal Generality (M-Gen.), Text Locality (T-Loc.), and Multimodal Locality (M-Loc.) after 1, 100, and 1000 sequential edits.}
\label{tab:main}
\renewcommand{\arraystretch}{1.30}
    \resizebox{\textwidth}{!}{ 
\begin{tabular}{ccccccccccccccccc}
\toprule
\multirow{3}{*}{Edits}  & \multicolumn{1}{c}{\multirow{3}{*}{Methods}} & \multicolumn{5}{c}{VLKEB}                                                                                    & \multicolumn{5}{c}{E-IC}                                                                                    & \multicolumn{5}{c}{E-VQA}                                                                                   \\ \cmidrule(lr){3-7} \cmidrule(lr){8-12} \cmidrule(lr){13-17} 
                        & \multicolumn{1}{c}{}                         & \multicolumn{1}{c}{\multirow{2}{*}{Reliability}} & \multicolumn{2}{c}{Generality}  & \multicolumn{2}{c}{Locality}   & \multicolumn{1}{c}{\multirow{2}{*}{Reliability}} & \multicolumn{2}{c}{Generality}  & \multicolumn{2}{c}{Locality}  & \multicolumn{1}{c}{\multirow{2}{*}{Reliability}} & \multicolumn{2}{c}{Generality}  & \multicolumn{2}{c}{Locality}  \\
                        & \multicolumn{1}{c}{}                         & \multicolumn{1}{c}{}                      & T-Gen.         & M-Gen.         & T-Loc.          & M-Loc.       & \multicolumn{1}{c}{}                      & T-Gen.         & M-Gen.         & T-Loc.       & M-Loc.         & \multicolumn{1}{c}{}                      & T-Gen.         & M-Gen.         & T-Loc.       & M-Loc.         \\ \hline
\multirow{8}{*}{T=1}    & FT-L                                         & 94.29                                     & 87.00             & 92.22          & 91.16           & 91.37        & 73.48                                     & 72.98          & 65.79          & 99.28        & 99.06          & 93.88                                     & 87.98          & 80.25          & 99.61        & 94.78          \\
                        & FT-M                                         & 76.31                                     & 65.57          & 59.43          & 100.00             & 92.35        & 56.19                                     & 56.55          & 49.94          & 100.00          & 100.00            & 87.29                                     & 76.11          & 53.23          & 100.00          & 96.95          \\
                        & MEND                                         & 92.13                                     & 91.28          & 90.22          & 89.19           & 90.13        & 92.82                                     & 91.81          & 90.59          & 96.38        & 93.69          & 91.23                                     & 90.05          & 91.29          & 91.02        & 90.22          \\
                        & SERAC                                        & 89.77                                     & 89.11          & 87.92          & 66.68           & 14.20         & 88.18                                     & 81.03          & 85.61          & 84.01        & 28.58          & 89.33                                     & 83.72          & 84.97          & 82.05        & 23.78          \\
                        & LEMoE                                        & 94.85                                     & 93.09          & 91.67          & 87.03           & 87.88        & 93.80                                      & 91.42          & 90.61          & 95.14        & 93.00             & 93.60                                      & 92.77          & 89.99          & 99.28        & 96.98          \\
                        & RECIPE                                       & 92.67                                     & 92.35          & 91.01          & 89.67           & 82.85        & 84.45                                     & 76.97          & 81.57          & 96.53        & 96.37          & 91.37                                     & 86.51          & 87.73          & 94.27        & 88.88          \\
                        & LiveEdit                                     & 96.43                                     & 95.22          & \textbf{93.72} & 100.00             & \textbf{100.00} & 82.16                                     & 81.01          & 78.27          & 100.00          & \textbf{100.00}            & 94.28                                     & \textbf{94.51}          & 88.01          & 100.00          & 100.00            \\ \cline{2-17} 
                        \addlinespace[1pt] 
                        & \gc ROSE                                         & \gc\textbf{100.00}                           & \gc \textbf{98.85} & \gc 88.87          & \gc \textbf{100.00} & \gc 98.72        & \gc \textbf{100.00}                                        & \gc \textbf{94.23}             & \gc \textbf{94.90}             & \gc \textbf{100.00}           & \gc 99.55             & \gc \textbf{100.00}                                        & \gc 93.92             & \gc \textbf{92.92}             & \gc \textbf{100.00}           & \gc \textbf{100.00}    \\     
                        \addlinespace[1pt]    
                        \hline
\multirow{8}{*}{T=100}  & FT-L                                         & 75.41                                     & 73.67          & 74.16          & 70.01           & 82.05        & 65.08                                     & 60.90           & 58.46          & 86.82        & 89.19          & 79.67                                     & 70.05          & 64.07          & 83.47        & 54.44          \\
                        & FT-M                                         & 60.60                                      & 59.79          & 56.29          & 100.00             & 68.07        & 59.41                                     & 55.44          & 52.16          & 100.00          & 72.93          & 82.90                                      & 72.83          & 47.26          & 100.00          & 43.39          \\
                        & MEND                                         & 0.56                                      & 0.58           & 0.66           & 0.18            & 0.07         & 56.83                                     & 56.89          & 53.07          & 87.63        & 84.64          & 2.22                                      & 2.20            & 2.21           & 0.21         & 0.62           \\
                        & SERAC                                        & 72.25                                     & 62.43          & 70.68          & 53.73           & 13.69        & 53.35                                     & 53.70           & 49.41          & 48.04        & 17.25          & 88.08                                     & 81.53          & 82.48          & 62.13        & 12.90           \\
                        & LEMoE                                        & 83.07                                     & 75.36          & 71.72          & 54.09           & 49.68        & 55.36                                     & 53.14          & 51.12          & 87.77        & 81.35          & 42.41                                     & 36.60           & 34.33          & 78.57        & 53.28          \\
                        & RECIPE                                       & 73.97                                     & 63.73          & 72.69          & 86.2            & 82.59        & 53.23                                     & 53.75          & 49.36          & 87.64        & 95.52          & 89.86                                     & 83.32          & 84.82          & 87.37        & 85.08          \\
                        & LiveEdit                                     & 94.56                                     & 90.65          & 89.56          & 100.00             & \textbf{100.00} & 80.81                                     & 78.77          & 63.52          & 100.00          & 100.00            & 93.54                                     & \textbf{92.34} & \textbf{85.89} & 100.00          & 99.31          \\ \cline{2-17} 
                        \addlinespace[1pt]
                        & \gc ROSE                                         & \gc \textbf{98.22}                            & \gc \textbf{95.00}    & \gc \textbf{93.80}  & \gc \textbf{100.00}    & \gc 95.31        & \gc \textbf{93.28}                            & \gc \textbf{87.55} & \gc \textbf{82.30}  & \gc \textbf{100.00} & \gc \textbf{100.00}   & \gc \textbf{97.00}                               & \gc 85.22          & \gc 83.97          & \gc \textbf{100.00} & \gc \textbf{99.38} \\ 
                        \addlinespace[1pt]
                        \hline
\multirow{8}{*}{T=1000} & FT-L                                         & 68.14                                     & 66.38          & 66.98          & 65.61           & 75.35        & 59.78                                     & 54.99          & 54.17          & 65.37        & 78.96          & 71.39                                     & 59.83          & 57.41          & 55.55        & 48.99          \\
                        & FT-M                                         & 53.41                                     & 48.80           & 43.16          & 100.00             & 57.03        & 49.21                                     & 47.75          & 43.81          & 100.00          & 35.14          & 69.57                                     & 56.34          & 44.07          & 100.00          & 41.47          \\
                        & MEND                                         & 0.03                                      & 0.05           & 0.07           & 0.06            & 0.08         & 54.39                                     & 54.14          & 50.99          & 83.87        & 80.60           & 0.04                                      & 0.05           & 0.05           & 0.08         & 0.09           \\
                        & SERAC                                        & 60.93                                     & 56.49          & 60.06          & 52.94           & 15.04        & 52.93                                     & 53.44          & 49.01          & 49.91        & 16.65          & 85.57                                     & 75.58          & 82.01          & 62.46        & 15.69          \\
                        & LEMoE                                        & 67.97                                     & 61.07          & 58.16          & 48.48           & 44.06        & 34.50                                      & 31.38          & 28.14          & 82.09        & 75.88          & 30.80                                      & 25.75          & 24.32          & 71.45        & 46.23          \\
                        & RECIPE                                       & 62.00                                        & 56.84          & 61.50           & 85.37           & 82.07        & 53.11                                     & 53.48          & 48.99          & 87.93        & 94.84          & 87.00                                        & 76.81          & 83.09          & 86.95        & 87.03          \\
                        & LiveEdit                                     & 92.22                                     & 83.97          & 82.75          & 100.00             & \textbf{100.00} & 72.80                                      & 69.95          & 57.05          & 100.00          & \textbf{99.79} & 92.93                                     & \textbf{90.16} & \textbf{84.30}  & 100.00          & \textbf{96.43} \\ \cline{2-17} 
                        \addlinespace[1pt]
                        & \gc ROSE                                         & \gc \textbf{95.24}                            & \gc \textbf{92.74} & \gc \textbf{84.53} & \gc \textbf{100.00}    & \gc 88.39        & \gc \textbf{90.65}                            & \gc \textbf{82.52} & \gc \textbf{72.22} & \gc \textbf{100.00} & \gc 97.42          & \gc \textbf{97.59}                            & \gc 80.85          & \gc 77.81          & \gc \textbf{100.00} & \gc 95.52          \\ 
                        \addlinespace[1pt]
                        \hline
\end{tabular}
}
\end{table*}

\subsection{Main Results}
\label{sec:main_results}

\cref{tab:main} presents the performance on VLKEB and MMEdit under the multimodal lifelong editing setup. From these results, three key observations can be made.

\textbf{1. ROSE effectively mitigates catastrophic forgetting.} As shown in Table 1, while most baselines experience a sharp decline in Reliability as the number of edits increases (e.g., FT-L drops to $68.14\%$ on VLKEB at $T=1000$), ROSE maintains a superior $95.24\%$. From $T=1$ to $T=1000$, ROSE's VLKEB Reliability decreases by 4.76 percentage points, compared with 26.15 for FT-L. At $T=1000$, ROSE also exceeds LiveEdit in Reliability by 17.85 points on E-IC and 4.66 points on E-VQA, extending this advantage across both MMEdit tasks. This stability stems from our risk-guided orthogonal mechanism, which successfully navigates weight updates away from sensitive pre-trained subspaces while isolating new knowledge into orthogonal complements.

\textbf{2. Superior comprehension and generalization.} On VLKEB, ROSE exhibits robust performance in T-Gen and M-Gen, outperforming the strongest baseline, LiveEdit, by a significant margin at $T=1000$ (e.g., $92.74\%$ vs $83.97\%$ in T-Gen). This is achieved via our Perturbation-Guided Knowledge Subspace Fusion, which extracts and integrates only the salient subspaces encoding the new knowledge while maintaining the backbone's fundamental representational structure. This selective modification helps apply edited facts across diverse paraphrases and visual contexts, even when the input query substantially deviates from the original editing sample, while limiting changes to the model's built-in general capabilities.

\textbf{3. Outstanding locality preservation.} Maintaining the integrity of unrelated knowledge is a critical failure point for many editors (e.g., SERAC's M-Loc drops to $15.04\%$ on VLKEB). In contrast, ROSE achieves strong Locality ($100\%$ T-Loc and $88.39\%$ M-Loc). This is attributed to our coarse-to-fine integration strategy: the semantic alignment gating encourages irrelevant queries to bypass the auxiliary branch, while the perturbation-guided fusion executes granular modifications for relevant inputs, thereby limiting unintended factual interference.

Furthermore, consistent gains across LLaVA, MiniGPT-4, and Qwen2-VL (see Appendix \ref{sec:backbone_scale}) further validate that ROSE is architecture-agnostic.

\subsection{Ablation Studies}
We conduct extensive ablation studies on the VLKEB benchmark (Table \ref{tab:ablation}) to dissect the contribution of each core component in ROSE.

\begin{table}[t]
\caption{Ablation study of ROSE on the VLKEB benchmark.}
\label{tab:ablation}
\renewcommand{\arraystretch}{1.15}
    \resizebox{\columnwidth}{!}{ 
\begin{tabular}{ccccccc}
\toprule
\multirow{2}{*}{Settings} & \multirow{2}{*}{Reliability} & \multicolumn{2}{c}{Generality} & \multicolumn{2}{c}{Locality} & \multirow{2}{*}{Average} \\ 
                          &                              & T-Gen         & M-Gen.         & T-Loc.        & M-Loc.       &                          \\ \hline
Fine Tuning                        & 79.16                        & 76.75         & 75.94          & 14.54         & 7.44         & 50.77           \\
w/o Multimodal Decoupling & 91.05                        & 88.14         & 85.63          & 100.00        & 40.54        & 81.07           \\
w/o Knowledge Integration & 93.15                        & 90.81         & 90.05          & 57.38         & 15.83        & 69.44           \\
w/o Semantic Gating       & 91.98                        & 89.71         & 89.02          & 76.87         & 25.97        & 74.71           \\
w/o Knowledge Fusion      & 96.28                        & 93.90         & 83.30          & 100.00        & 84.37        & 91.57           \\
w/o Orthogonal Editing    & 78.71                        & 76.00         & 71.05          & 100.00        & 85.50        & 82.25           \\
w/o Risk Awareness        & 90.07                        & 87.31         & 78.09          & 100.00        & 88.14        & 88.72           \\ \hline
ROSE                      & 95.24                        & 92.74         & 84.53          & 100.00        & 88.39        & 92.18           \\ \hline
\end{tabular}
}
\end{table}

\textbf{Role of Risk-Aware Mask.} The mask $\mathcal{M}_{\text{risk}}$ in \cref{equ:mask-pert} is designed to protect pre-trained general knowledge by confining updates to active parameters of edited knowledge. As shown in \cref{tab:ablation}, omitting this mask results in a significant drop in both Reliability and Generality. In the absence of $\mathcal{M}_{\text{risk}}$, ROSE degenerates to enforcing orthogonal constraints over the entire parameter space. This over-constraint severely shrinks the feasible solution space, making it harder to converge to an optimal edit for new tasks. Furthermore, the results indicate that without parameter isolation, edits inadvertently impair the model's foundational representations, confirming that $\mathcal{M}_{\text{risk}}$ is essential for maintaining the balance between plasticity and stability.

\textbf{Effect of Orthogonal Lifelong Editing.} The orthogonal projection of gradients onto the complement of the subspace $Q^{(t)}$ is fundamental to ROSE's ability to perform lifelong editing. In \cref{tab:ablation}, ROSE achieves $95.24\%$ Reliability versus $78.71\%$ without orthogonal editing, and $88.39\%$ M-Loc versus $40.54\%$ without multimodal decoupling. Removing the orthogonal constraint leads the model to suffer from immediate catastrophic forgetting, with its performance on earlier edits degrading quickly as more edits are applied. Additionally, removing multimodal decoupling introduces severe cross-modal entanglement, where textual updates disrupt visual representations (and vice versa). Together, these results underscore the complementary roles of orthogonal editing and independent handling of textual and visual subspaces in effective multimodal lifelong editing.

\textbf{Impact of Dual-Hierarchy Router.} ROSE balances plasticity and stability through two-tier filtering during inference. (1) The coarse-grained gate $\mathcal{G}$ prevents the auxiliary branch from being mistakenly activated by irrelevant inputs, thereby preserving model Locality. (2) The fine-grained, perturbation-based mask $\mathcal{M}_{\text{fusion}}$ ensures that only the salient knowledge subspaces are injected into the backbone's pretrained knowledge during fusion. \cref{tab:ablation} shows that omitting either component leads to incorrect routing or ``knowledge leakage,'' where edited facts interfere with general queries. This validates that a hierarchical routing mechanism is necessary to achieve high-precision knowledge retrieval in complex multimodal scenarios.

\subsection{Further Studies}\label{sec:fs}

\begin{table}[htbp]
\centering
\renewcommand{\arraystretch}{0.9} %

\begin{minipage}[t]{0.49\textwidth}
\centering
\caption{Impact of the risk sensitivity threshold $k$.}
\label{tab:risk}
\renewcommand{\arraystretch}{1.3}
\resizebox{\linewidth}{!}{ 
\begin{tabular}{ccccccc}
\toprule
\multirow{2}{*}{Settings} & \multirow{2}{*}{Reliability} & \multicolumn{2}{c}{Generality} & \multicolumn{2}{c}{Locality} & \multirow{2}{*}{Average} \\
                          &                              & T-Gen.         & M-Gen.        & T-Loc.        & M-Loc.       &                          \\ \hline
0.1                       & 71.42                        & 63.63          & 63.53         & 100.00        & 89.86        & 77.69           \\
0.3                       & 91.02                        & 79.28          & 77.23         & 100.00        & 89.44        & 87.39           \\
0.5                       & 95.24                        & 92.74          & 84.53         & 100.00        & 88.39        & 92.18           \\
0.7                       & 95.23                        & 91.63          & 82.10         & 100.00        & 88.59        & 91.51           \\
0.9                       & 96.55                        & 92.88          & 83.38         & 100.00        & 88.27        & 92.22           \\
1.0                         & 90.07                        & 87.31          & 78.09         & 100.00        & 88.14        & 88.72           \\ \hline
\end{tabular}
}
\end{minipage}\hfill
\begin{minipage}[t]{0.49\textwidth}
\centering 
\caption{Impact of the perturbation-guided fusion threshold $\sigma$.}
\label{tab:fusionGate}
\renewcommand{\arraystretch}{1.3}
\resizebox{\linewidth}{!}{ 
\begin{tabular}{cccccccc}
\toprule
\multirow{2}{*}{Edits}  & \multirow{2}{*}{Settings} & \multirow{2}{*}{Reliability} & \multicolumn{2}{c}{Generality} & \multicolumn{2}{c}{Locality} & \multirow{2}{*}{Average} \\
                        &                           &                              & T-Gen.         & M-Gen         & T-Loc.        & M-Loc.       &                          \\ \hline      
\multirow{3}{*}{T=1}    & 1                         & 100.00                       & 99.10          & 88.87         & 100.00        & 97.89        & 97.17                    \\
                        & 2                         & 100.00                       & 98.85          & 88.87         & 100.00        & 98.72        & 97.29                    \\
                        & 3                         & 100.00                       & 98.15          & 88.87         & 100.00        & 99.26        & 97.26                    \\ \hline
\multirow{3}{*}{T=100}  & 1                         & 98.00                        & 93.77          & 89.55         & 100.00        & 94.91        & 95.25                    \\
                        & 2                         & 98.22                        & 95.00          & 93.80         & 100.00        & 95.31        & 96.47                    \\
                        & 3                         & 98.01                        & 92.40          & 89.08         & 100.00        & 97.23        & 95.34                    \\ \hline
\multirow{3}{*}{T=1000} & 1                         & 94.52                        & 91.30          & 81.57         & 100.00        & 86.30        & 90.74                    \\
                        & 2                         & 95.24                        & 92.74          & 84.53         & 100.00        & 88.39        & 92.18                    \\
                        & 3                         & 93.10                        & 85.65          & 79.52         & 100.00        & 92.76        & 90.21                    \\ \hline 
\end{tabular}
}
\end{minipage}
\end{table}

\textbf{Impact of the risk sensitivity threshold $k$.} The quantile $k$ in $\mathcal{M}_{\text{risk}}$ serves as a threshold that controls how sensitively risks are detected. As shown in \cref{tab:risk}, a small $k$ (e.g., $0.1$) adopts a conservative strategy, targeting only those subspaces exhibiting the most pronounced perturbations, which severely limits the acquisition of new knowledge and drops Reliability. Conversely, an overly large $k$ (e.g., $0.9$) places ROSE in a highly sensitive regime, in which a broader range of parameter perturbations is labeled as risky. This enlarges the solution space but unintentionally encroaches upon general-purpose subspaces, weakening Locality. We found that $k=0.5$ strikes a reasonable balance, enabling ROSE to accurately localize risky subspaces while limiting erosion of foundational capabilities.

\begin{figure}[htbp]
    \centering
    \begin{minipage}[t]{0.49\textwidth}
        \centering
        \includegraphics[width=\linewidth]{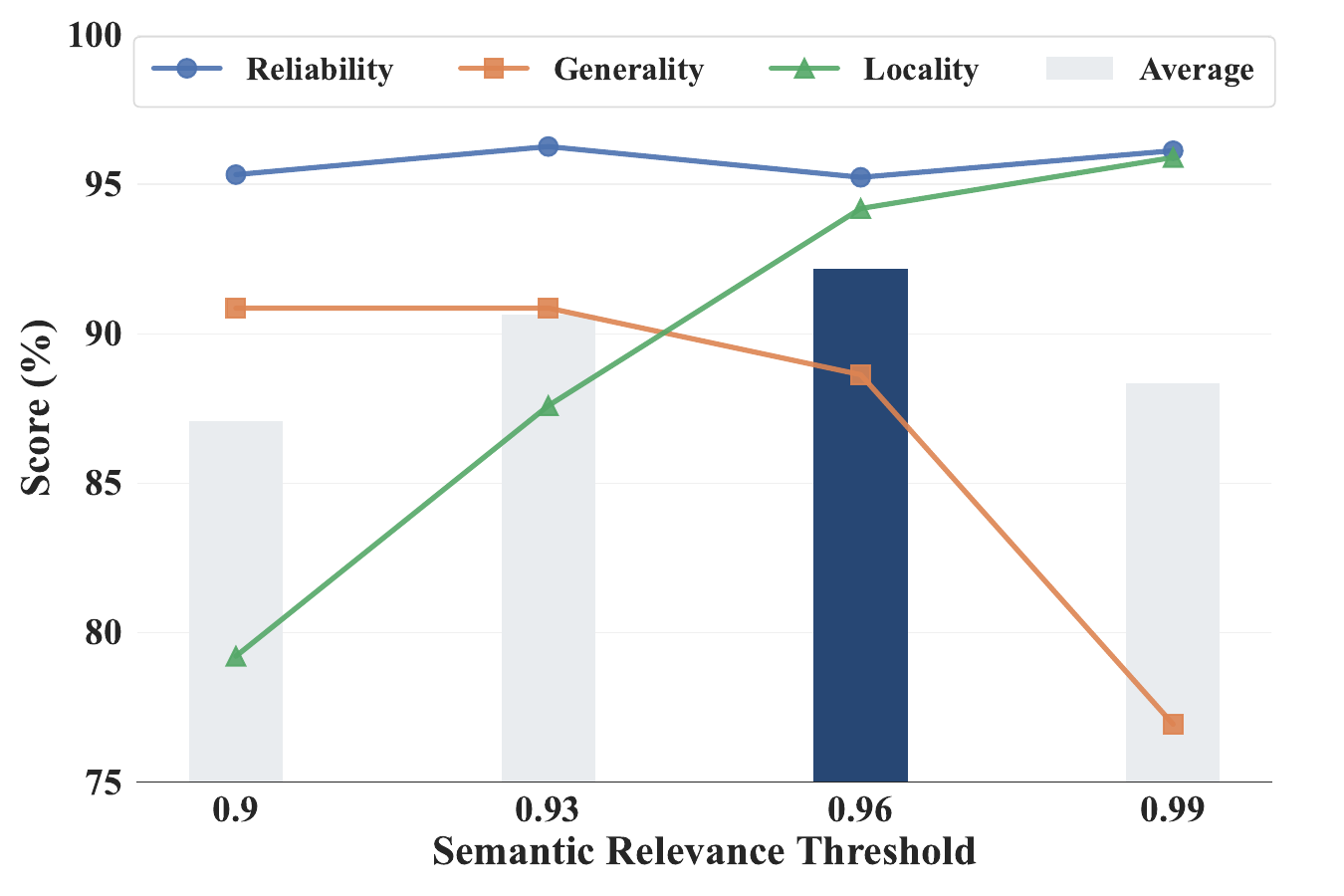} 
        \caption{Sensitivity analysis of the semantic gating threshold $\tau$. The ``Average" represents the mean across all metrics.}%
        \label{fig:semanticGate}
    \end{minipage}
    \hfill %
    \begin{minipage}[t]{0.49\textwidth}
        \centering
        \includegraphics[width=\linewidth]{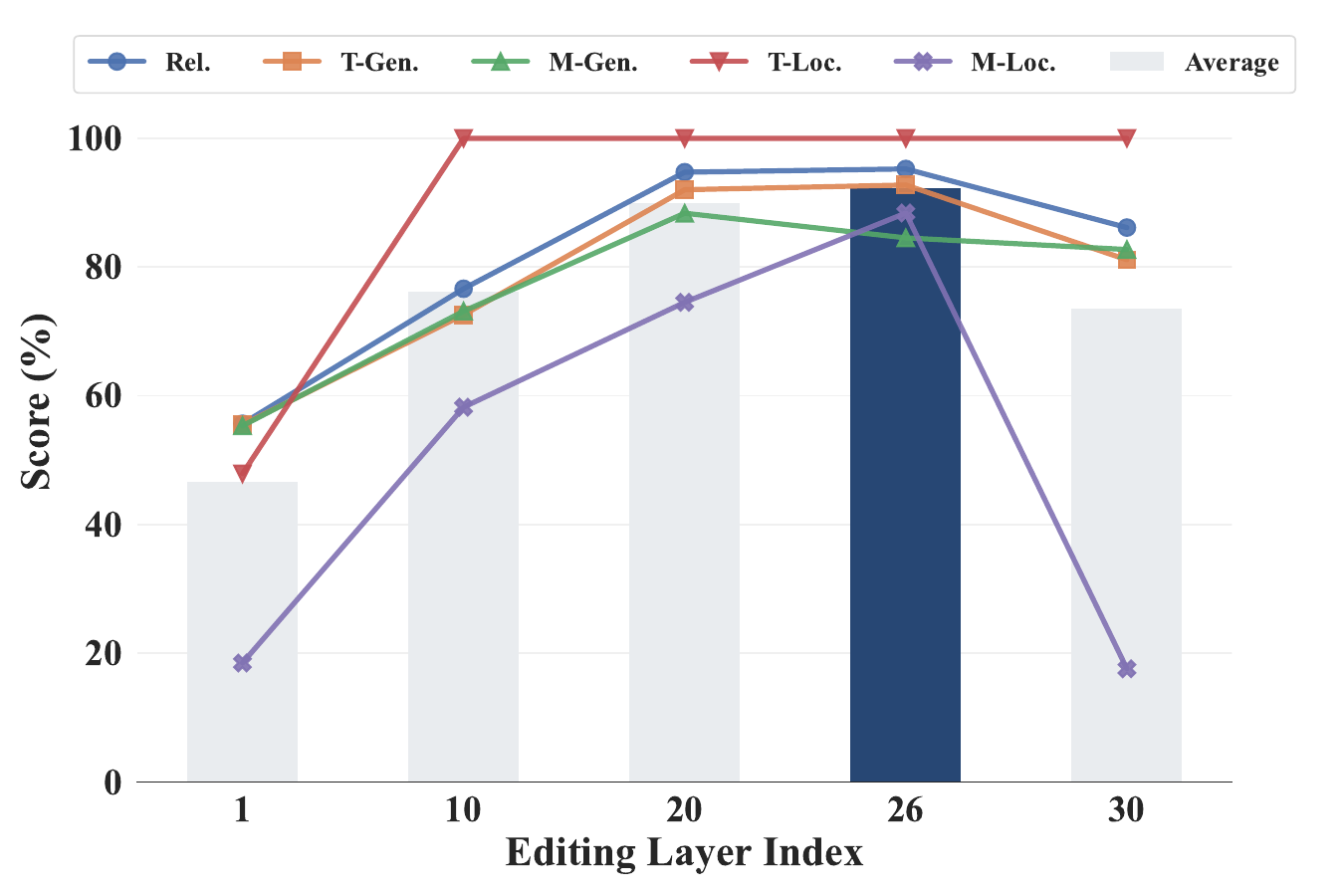} 
        \caption{Effect of the attached layer index selected for editing. The ``Average" denotes the mean value of all metrics. }%
        \label{fig:attached_layer}
    \end{minipage}
\end{figure}

\textbf{Analysis of the Semantic Gating.} We examine how the gating threshold $\tau$ affects performance (\cref{fig:semanticGate}), keeping $\tau_{\text{min}}=0.4$ and $\tau_{\text{gap}}=0.01$ fixed. While Reliability remains robust across a wide range of thresholds, Generality and Locality exhibit an intrinsically adversarial relationship. A low $\tau$ yields an overly permissive router that mistakenly triggers the editing branch for unrelated inputs, harming Locality. Conversely, a high $\tau$ improves Locality but struggles to recognize semantically rephrased queries, reducing Generality. The apparent sharp peak in the average metric reflects the balance between these two competing objectives. At $\tau = 0.96$, the router achieves a balance by accurately triggering edits for pertinent inputs while remaining safely inactive for the irrelevant distribution, thereby maximizing the model's overall stability.

\textbf{Analysis of the saliency of Knowledge Fusion.} The threshold $\sigma$ controls the intensity of edited knowledge fusion (\cref{tab:fusionGate}). A small $\sigma$ leads to excessive mixing, injecting auxiliary feature into insignificant subspaces and causing interference. This unnecessary interference erodes pretrained knowledge and undermines locality. In contrast, a large $\sigma$ causes under-mixing by confining fusion to only the strongest activation peaks. This suppresses more nuanced contextual dependencies, leading to incomplete semantics and reduced reliability. Setting $\sigma=2$ yields highly targeted integration, capturing key novel semantics while maintaining the backbone's structure within invariant subspaces.

\textbf{Effect of Editing Depth.} We analyze how ROSE performs when integrated at different network depths. As shown in \cref{fig:attached_layer}, adding ROSE to early layers (e.g., 1-10) harms performance because these layers primarily encode low-level perceptual details rather than abstract factual associations. Similarly, final layers contain highly compressed, task-specific representations that lack the semantic breadth needed for editing. The best results emerge when ROSE is applied to mid-to-late layers (e.g., Layer 26), indicating that factual knowledge is concentrated in an intermediate semantic space where strong cross-modal alignment supports accurate editing.

\textbf{Effect of the Reference Anchor.} The comparison in Appendix \ref{app:more_exp} shows better Locality preservation with the static backbone anchor $\boldsymbol{W}_{\text{down}}$ than with the sequential anchor $\boldsymbol{W}_{t-1}$. A fixed reference keeps risk identification tied to the original pre-trained weights throughout the editing sequence, supporting ROSE's use of a consistent reference for protecting foundational knowledge.

\subsection{Analysis of Computational Cost}
\label{sec:comp_cost}

In lifelong knowledge editing over continuous and long data streams, computational efficiency is paramount. To evaluate the operational overhead of the multiple components within our framework (such as orthogonal projection and multi-granularity routing), we measured the average editing time, inference time, and peak memory consumption of ROSE. All efficiency evaluations were conducted on a single NVIDIA RTX 4090 GPU using the LLaVA-v1.5-7B backbone. The results are summarized in \cref{tab:comp_cost}.

\begin{table}[ht]
\centering
\caption{Computational cost of ROSE across different numbers of sequential edits. Metrics include average editing time per edit, average inference time per query, and peak VRAM consumption.}
\label{tab:comp_cost}
\renewcommand{\arraystretch}{1.15}
\begin{tabular}{cccc}
\toprule
\textbf{Edits} & \textbf{Editing Time (s/edit)} & \textbf{Inference Time (s/query)} & \textbf{Peak VRAM (MB)} \\
\midrule
\textbf{1}    & 1.6748 & 0.9973 & 16781.52 \\
\textbf{10}   & 1.7113 & 1.1849 & 16826.89 \\
\textbf{100}  & 1.3975 & 1.1248 & 16857.73 \\
\textbf{1000} & 1.3140 & 1.1256 & 17005.69 \\
\bottomrule
\end{tabular}
\end{table}

As shown in \cref{tab:comp_cost}, the \textbf{Editing Time} remains highly stable at approximately $1.3$ to $1.7$ seconds per edit, even as the sequence scales up to $1,000$ edits. This indicates that the orthogonal projection mechanism does not incur escalating computational burdens over time. Similarly, the \textbf{Inference Time} is highly consistent at around $1.1$ seconds per query. This demonstrates that the proposed multi-granularity semantic gating and perturbation-driven fusion mechanisms introduce virtually no noticeable latency during the inference phase. 

Furthermore, the \textbf{Peak VRAM} usage grows extremely slowly. Scaling from a single edit to $1,000$ continuous edits adds only about $224$ MB of memory overhead. This confirms that our explicit parameter isolation strategy, utilizing auxiliary branches and maintaining compact orthogonal matrices, is highly memory-efficient, making ROSE practically viable for long-term lifelong editing scenarios.

For extreme lifelong scenarios where the accumulated feature subspace approaches full rank, we discuss a possible Singular Value Decomposition (SVD)-based basis truncation strategy (see Appendix \ref{sec:theoretical_discussions}) to manage knowledge decay gracefully.

\section{Conclusion}

In this paper, we present ROSE, a lifelong knowledge editing framework for MLLMs that strikes a balance between plasticity and stability. By aiming to restrict parameter updates to the orthogonal complement of the edited knowledge subspace and applying a risk mask, ROSE navigates the knowledge space and reduces catastrophic interference between sequential edits. Moreover, our multi-granularity integration scheme enables precise activation of the relevant edited knowledge while limiting locality erosion. Extensive experiments demonstrate that ROSE consistently outperforms existing baselines.

\clearpage
\begin{ack}
This work was supported by the National Natural Science Foundation of China (62576283, 62402390, 62607058, 62672417, 62302397), Natural Science Basic Research Program of Shaanxi (2026JC-YXQN-252, 2026JC-YXQN-234, 2026JC-YBQN-0841), Guangdong Basic and Applied Basic Research Foundation under Grant No. 2025A1515012995, Young Talent Fund of Xi'an Association for Science and Technology (0959202613046), Young Elite Scientists Sponsorship Program by CAST (YESS20240039), Postdoctoral Research Project Funding of Shaanxi Province (2025BSHTDZZ005, 2025BSHSDZZ106), China Postdoctoral Science Foundation (2025M774364), and Open Project Program of the State Key Laboratory of Medical Proteomics (SKLP-O202514).
\end{ack}

\medskip
\small
\bibliographystyle{unsrtnat} 
\bibliography{ref}
\normalsize

\newpage
\appendix
\onecolumn
\section{Detailed Evaluation Metrics}
\label{app:metrics}

In this appendix, we provide the formal definitions of the evaluation metrics used in knowledge editing: reliability, generality, and locality.

\paragraph{Reliability.} 
Reliability quantifies the success rate of the edited model on the exact instances it was modified to handle. For a given edit $e = (v_e, p_e, o^*_e)$, where $o^*_e$ represents the target output, reliability is defined as:
\begin{equation}
    \text{Reliability} = \mathbb{E}_{e \sim \mathcal{D}_{\text{edit}}} \, \mathbb{I} \left\{ f(v_e, p_e; \theta') = o^*_e \right\},
\end{equation}
where $\mathbb{I}\{\cdot\}$ denotes the indicator function, and $\theta'$ represents the parameters of the edited model.

\paragraph{Generality.} 
Generality measures the model's ability to apply the edited knowledge to semantically equivalent samples within a neighborhood $\mathcal{N}(e)$. These samples typically include rephrased textual prompts $p_{\text{re}}$ or visually transformed images $v_{\text{re}}$ that convey the same underlying fact as the original edit. The metric is formulated as:
\begin{equation}
    \text{Generality} = \mathbb{E}_{e \sim \mathcal{D}_{\text{edit}}} \, \mathbb{E}_{(v_{\text{re}}, p_{\text{re}}) \sim \mathcal{N}(e)} \, \mathbb{I} \left\{ f(v_{\text{re}}, p_{\text{re}}; \theta') = o^*_e \right\}.
\end{equation}

\paragraph{Locality.} 
Locality evaluates whether the editing process unintentionally alters the model's behavior on unrelated knowledge. Let $\mathcal{D}_{\text{loc}}$ denote a set of irrelevant samples $(v_{\text{loc}}, p_{\text{loc}})$ that are independent of the edit $e$. Locality quantifies the extent to which the original pretrained knowledge is preserved:
\begin{equation}
    \text{Locality} = \mathbb{E}_{(v_{\text{loc}}, p_{\text{loc}}) \sim \mathcal{D}_{\text{loc}}} \, \mathbb{I} \left\{ f(v_{\text{loc}}, p_{\text{loc}}; \theta') = f(v_{\text{loc}}, p_{\text{loc}}; \theta) \right\},
\end{equation}
where $\theta$ represents the parameters of the original pre-trained model.

\section{Mathematical Derivations and Proofs}
\label{app:proof_omission}

In this appendix, we establish the formal mathematical foundation for analyzing representation-level sensitivity. This analysis directly inspires the perturbation-driven fusion mask utilized in our multi-granularity knowledge integration strategy.

Recent evidence \cite{shenfeld2025rl} links distributional shifts to catastrophic forgetting during fine-tuning. Building on this insight, Theorem~\ref{thm:KL} derives a first-order approximation of the fixed-target KL loss change caused by an intermediate perturbation under the linear mapping defined below, and Corollary~\ref{cor:kl_bound} bounds this change by a first-order contribution and a controlled remainder, thereby providing theoretical motivation for the perturbation-driven fusion mask in Section~\ref{sec:module2}, where $|\Delta\boldsymbol{h}_{i,j}|$ serves as a computationally feasible surrogate for identifying actively modified feature coordinates. We first introduce the necessary definitions and lemmas, then present the proof of the theorem and its corollary.

\subsection{Preliminaries}
\label{app:def}
\begin{definition}[Logit Mapping and Probability]
To theoretically analyze representation-level sensitivity at an intermediate layer $l$, we simulate mapping its output representation $\boldsymbol{h} \in \mathbb{R}^{d_m}$ for a single token directly to the vocabulary space. Specifically, we obtain the logits $\boldsymbol{z} \in \mathbb{R}^D$ via a linear projection $\boldsymbol{z} = \boldsymbol{W}_l^\top \boldsymbol{h}$, where $\mathbf{W}_l \in \mathbb{R}^{d_m \times D}$. The predictive distribution $\boldsymbol{q} \in \mathbb{R}^{D}$ is then given by the softmax function:
\begin{equation}
    \boldsymbol{q}_k(\mathbf{z}) = \frac{\text{exp}(\boldsymbol{z}_k)}{\sum_{j=1}^D \text{exp}
    (\boldsymbol{z}_j)}, \quad \forall k \in \{1, \dots, D\}.
\end{equation}
\end{definition}

\begin{definition}[Objective Function]
Let $\boldsymbol{p} \in \mathbb{R}^{D}$ be the fixed target distribution, with $p_k \ge 0$ and $\sum_k p_k=1$. We use the convention $0\log 0=0$ and minimize the Kullback-Leibler divergence $\mathbb{D}_{\text{KL}}(\boldsymbol{p} \| \boldsymbol{q})$:
\begin{equation}
    \mathcal{L}(\mathbf{\boldsymbol{z}}) = \mathbb{D}_{\text{KL}}(\boldsymbol{p} \| \boldsymbol{q}) = -\sum_{k=1}^D \boldsymbol{p}_k \text{log} \boldsymbol{q}_k(\boldsymbol{z}) + \sum_{k=1}^D \boldsymbol{p}_k \log \boldsymbol{p}_k.
\end{equation}
\end{definition}

To analyze the sensitivity of the objective, we first require the Jacobian of the softmax function and the gradient of the loss with respect to the logits.

\begin{lemma}[Jacobian of Softmax]
\label{lemma:softmax_jacobian}
The partial derivative of the softmax output $q_k$ with respect to the logit $z_i$ is:
\begin{equation}
    \frac{\partial q_k}{\partial z_i} = q_k (\delta_{ki} - q_i),
\end{equation}
where $\delta_{ki}$ is the Kronecker delta, which is 1 if $k=i$ and 0 otherwise.
\end{lemma}

\begin{proof}
Applying the quotient rule to $q_k$:
\begin{align*}
    \frac{\partial q_k}{\partial z_i} &= \frac{\partial}{\partial z_i}\left(\frac{e^{z_k}}{\sum_j e^{z_j}}\right) \\
    &= \frac{\frac{\partial e^{z_k}}{\partial z_i} (\sum_j e^{z_j}) - e^{z_k} \frac{\partial (\sum_j e^{z_j})}{\partial z_i}}{(\sum_j e^{z_j})^2} \\
    &= \frac{e^{z_k}\delta_{ki}(\sum_j e^{z_j}) - e^{z_k} e^{z_i}}{(\sum_j e^{z_j})^2} \\
    &= \frac{e^{z_k}}{\sum_j e^{z_j}}\delta_{ki} - \frac{e^{z_k}}{\sum_j e^{z_j}}\frac{e^{z_i}}{\sum_j e^{z_j}} \\
    &= q_k \delta_{ki} - q_k q_i \\
    &= q_k (\delta_{ki} - q_i).
\end{align*}
\end{proof}

\begin{lemma}[Gradient of KL Divergence]
\label{lemma:kl_gradient}
The gradient of the objective function $\mathcal{L}$ with respect to the logit vector $\mathbf{z}$ is:
\begin{equation}
    \nabla_{\mathbf{z}} \mathcal{L} = \boldsymbol{q} - \boldsymbol{p}.
\end{equation}
\end{lemma}

\begin{proof}
Using the chain rule and Lemma \ref{lemma:softmax_jacobian}:
\begin{align*}
    \frac{\partial \mathcal{L}}{\partial z_i} &= -\sum_{k=1}^D p_k \frac{\partial \log q_k}{\partial z_i} = -\sum_{k=1}^D \frac{p_k}{q_k} \frac{\partial q_k}{\partial z_i} \\
    &= -\sum_{k=1}^D \frac{p_k}{q_k} q_k (\delta_{ki} - q_i) \\
    &= -\sum_{k=1}^D p_k (\delta_{ki} - q_i) \\
    &= -\left( \sum_{k=1}^D p_k \delta_{ki} - \sum_{k=1}^D p_k q_i \right) \\
    &= -(p_i - q_i \sum_{k=1}^D p_k).
\end{align*}
Since $\sum_{k=1}^D p_k = 1$, we have $\frac{\partial \mathcal{L}}{\partial z_i} = q_i - p_i$. Thus, $\nabla_{\mathbf{z}} \mathcal{L} = \boldsymbol{q} - \boldsymbol{p}$.
\end{proof}

With these preliminary results in place, we now formally state and prove the main sensitivity theorem.

\subsection{The Sensitivity of Knowledge Perturbation}
\label{app:proof_theorem}

\begin{theorem}[The Sensitivity of Knowledge Perturbation]
\label{thm:KL}
Consider an intermediate representation $\boldsymbol{h} \in \mathbb{R}^{d_m}$ at layer $l$. Let $\boldsymbol{W}_{l} \in \mathbb{R}^{d_m \times D}$ map it to logits $\boldsymbol{z}\in \mathbb{R}^D$ via $\boldsymbol{z} = \boldsymbol{W}_{l}^\top\boldsymbol{h}$. With the KL divergence $\mathbb{D}_{\text{KL}}$, a target distribution $\boldsymbol{p}\in\mathbb{R}^D$, and a predictive distribution $\boldsymbol{q} = \sigma(\boldsymbol{z})\in\mathbb{R}^D$ obtained via the softmax function $\sigma$, the sensitivity of $\mathbb{D}_{\text{KL}}(\boldsymbol{p}\Vert\boldsymbol{q})$ to a small intermediate perturbation $\boldsymbol{\delta}_l \in \mathbb{R}^{d_m}$ at layer $l$ can be written as:
\begin{equation}
\label{equ:DKL}
    \Delta \mathbb{D}_{\text{KL}}(\boldsymbol{\delta}_l) = (\boldsymbol{q} - \boldsymbol{p})^\top \boldsymbol{W}_{l}^\top \boldsymbol{\delta}_l + \mathcal{O}(\|\boldsymbol{\delta}_l\|^2),
\end{equation}
where $\Delta\mathbb{D}_{\text{KL}}(\boldsymbol{\delta}_l)=\mathcal{L}(\boldsymbol{z}+\boldsymbol{W}_l^\top\boldsymbol{\delta}_l)-\mathcal{L}(\boldsymbol{z})$, with $\boldsymbol{p}$ and $\boldsymbol{W}_l$ held fixed, and $\mathcal{O}(\|\boldsymbol{\delta}_l\|_2^2)$ encapsulates the higher-order terms.
\end{theorem}

\begin{proof}
Let $\mathbf{h} \in \mathbb{R}^{d_m}$ denote the latent representation at layer $l$, and let $\boldsymbol{\delta}_l \in \mathbb{R}^{d_m}$ be a small additive perturbation. The mapping from the latent space to the logits $\mathbf{z} \in \mathbb{R}^D$ is defined by a linear transformation $\mathbf{W}_l \in \mathbb{R}^{d_m \times D}$, such that $\mathbf{z} = \mathbf{W}_l^\top \mathbf{h}$.
Consequently, the perturbation induced in the logit space, denoted as $\mathbf{v} \in \mathbb{R}^D$, is given by:
\begin{equation}
    \mathbf{v} = \Delta \mathbf{z} = \mathbf{W}_l^\top (\mathbf{h} + \boldsymbol{\delta}_l) - \mathbf{W}_l^\top \mathbf{h} = \mathbf{W}_l^\top \boldsymbol{\delta}_l.
\end{equation}

The objective function is the KL divergence $\mathcal{L}(\mathbf{z}) = \mathbb{D}_{\mathrm{KL}}(\mathbf{p} \| \mathbf{q}(\mathbf{z}))$. Our goal is to analyze the sensitivity variation $\Delta \mathbb{D}_{\mathrm{KL}}(\boldsymbol{\delta}_l) = \mathcal{L}(\mathbf{z} + \mathbf{v}) - \mathcal{L}(\mathbf{z})$.

We construct a parameterized scalar function $g(t) = \mathcal{L}(\mathbf{z} + t\mathbf{v})$ for $t \in [0, 1]$. For $t\in(0,1]$, Taylor's theorem with Lagrange remainder around $t=0$ gives:
\begin{equation}
    g(t) = g(0) + g'(0)t + \frac12 g''(\xi_t)t^2,\qquad 0<\xi_t<t,
\end{equation}
where $g''(s)=\mathbf{v}^\top H(\mathbf{z}+s\mathbf{v})\mathbf{v}$ and $H(\mathbf{z})=\nabla^2\mathcal{L}(\mathbf{z})=\operatorname{diag}(\boldsymbol{q}(\mathbf{z}))-\boldsymbol{q}(\mathbf{z})\boldsymbol{q}(\mathbf{z})^\top$. At $t=0$, the equality holds directly.

According to Taylor's Theorem, the total variation is:
\begin{equation}
    \Delta \mathbb{D}_{\mathrm{KL}}(\boldsymbol{\delta}_l) = g(1) - g(0) = g'(0) + R_1(\boldsymbol{\delta}_l),
\end{equation}
where $R_1(\boldsymbol{\delta}_l)=\frac12\mathbf{v}^\top H(\mathbf{z}+\xi\mathbf{v})\mathbf{v}$ for some $\xi\in(0,1)$. The Hessian $H$ is positive semidefinite because its quadratic form is a variance under $\boldsymbol{q}$. Its maximum absolute row sum is $\max_i 2q_i(1-q_i)\le 1/2$; since $H$ is symmetric, $\|H\|_2\le 1/2$. Hence,
\begin{equation}
\label{eq:kl_remainder}
    0\le R_1(\boldsymbol{\delta}_l)\le\frac14\|\mathbf{v}\|_2^2=\frac14\|\mathbf{W}_l^\top\boldsymbol{\delta}_l\|_2^2.
\end{equation}

To find the linear term $g'(0)$, we apply the multivariate Chain Rule:
\begin{equation}
    g'(t) = \nabla_{\mathbf{z}} \mathcal{L}(\mathbf{z} + t\mathbf{v})^\top \frac{\partial}{\partial t}(\mathbf{z} + t\mathbf{v}) = \nabla_{\mathbf{z}} \mathcal{L}(\mathbf{z} + t\mathbf{v})^\top \mathbf{v}.
\end{equation}
Evaluating at $t=0$, we have $g'(0) = \nabla_{\mathbf{z}} \mathcal{L}(\mathbf{z})^\top \mathbf{v}$.

By Lemma \ref{lemma:kl_gradient}, the gradient of the KL divergence with respect to the logits is given explicitly by $\nabla_{\mathbf{z}} \mathcal{L} = \mathbf{q} - \mathbf{p}$. Substituting this result into the expression for $g'(0)$, we obtain:
\begin{equation}
    g'(0) = (\mathbf{q} - \mathbf{p})^\top \mathbf{v}.
\end{equation}

Finally, we map the sensitivity back to the intermediate perturbation $\boldsymbol{\delta}_l$ by substituting $\mathbf{v} = \mathbf{W}_l^\top \boldsymbol{\delta}_l$:
\begin{equation}
    \Delta \mathbb{D}_{\mathrm{KL}}(\boldsymbol{\delta}_l) = (\mathbf{q} - \mathbf{p})^\top (\mathbf{W}_l^\top \boldsymbol{\delta}_l) + R_1(\boldsymbol{\delta}_l).
\end{equation}
For fixed $\mathbf{W}_l$, \eqref{eq:kl_remainder} gives $R_1(\boldsymbol{\delta}_l)=\mathcal{O}(\|\boldsymbol{\delta}_l\|_2^2)$, which completes the proof.
\end{proof}

Building on Theorem~\ref{thm:KL}, we now derive a worst-case upper bound that isolates the role of the feature perturbation magnitude, which directly motivates the perturbation-driven fusion mask in \cref{sec:module2}.

\begin{corollary}[Worst-Case Bound for KL Loss Change]
\label{cor:kl_bound}
During inference, the true target distribution $\boldsymbol{p}$ is unknown, making it impossible to directly compute the first-order sensitivity term $(\boldsymbol{q} - \boldsymbol{p})^\top \boldsymbol{W}_{l}^\top \boldsymbol{\delta}_l$ in Theorem~\ref{thm:KL}. However, we can circumvent this difficulty by deriving a conservative upper bound on the fixed-target KL loss change that isolates the role of the feature perturbation $\boldsymbol{\delta}_l$. This bound provides a theoretically grounded yet practically feasible motivation for identifying the most actively modified feature dimensions without requiring knowledge of $\boldsymbol{p}$.

By applying H\"older's inequality to the first-order term in Theorem~\ref{thm:KL} and retaining the remainder bound \eqref{eq:kl_remainder}, we obtain:
\begin{equation}
\label{eq:kl_bound}
\begin{aligned}
    \left|\Delta \mathbb{D}_{\text{KL}}(\boldsymbol{\delta}_l)\right| 
    &\le \left| (\boldsymbol{q} - \boldsymbol{p})^\top \boldsymbol{W}_{l}^\top \boldsymbol{\delta}_l \right| + \frac14\|\boldsymbol{W}_l^\top\boldsymbol{\delta}_l\|_2^2 \\
    &\le \left\| \boldsymbol{W}_{l} (\boldsymbol{q} - \boldsymbol{p}) \right\|_1 \cdot \left\|\boldsymbol{\delta}_l\right\|_\infty + \frac14\|\boldsymbol{W}_l^\top\boldsymbol{\delta}_l\|_2^2.
\end{aligned}
\end{equation}
Alternatively, an element-wise decomposition provides a finer-grained perspective on how individual feature dimensions contribute to the worst-case bound:
\begin{equation}
\label{eq:element_bound}
    \left|\Delta \mathbb{D}_{\text{KL}}(\boldsymbol{\delta}_l)\right| 
    \le \sum_{i=1}^{d_m} \left| \bigl[ \boldsymbol{W}_{l} (\boldsymbol{q} - \boldsymbol{p}) \bigr]_i \right| \cdot \left| (\boldsymbol{\delta}_l)_i \right| + \frac14\|\boldsymbol{W}_l^\top\boldsymbol{\delta}_l\|_2^2.
\end{equation}
\end{corollary}

\begin{proof}
We begin from the exact decomposition established in the proof of Theorem~\ref{thm:KL}:
\begin{equation}
    \Delta \mathbb{D}_{\text{KL}}(\boldsymbol{\delta}_l) = (\boldsymbol{q} - \boldsymbol{p})^\top \boldsymbol{W}_{l}^\top \boldsymbol{\delta}_l + R_1(\boldsymbol{\delta}_l).
\end{equation}
Taking the absolute value and rewriting the inner product,
\begin{equation}
    \left|\Delta \mathbb{D}_{\text{KL}}(\boldsymbol{\delta}_l)\right|
    \le \left| (\boldsymbol{q} - \boldsymbol{p})^\top \boldsymbol{W}_{l}^\top \boldsymbol{\delta}_l \right| + |R_1(\boldsymbol{\delta}_l)|
    = \left| \bigl\langle \boldsymbol{W}_{l} (\boldsymbol{q} - \boldsymbol{p}),\, \boldsymbol{\delta}_l \bigr\rangle \right| + |R_1(\boldsymbol{\delta}_l)|,
\end{equation}
where $\langle\cdot,\cdot\rangle$ denotes the standard Euclidean inner product and we have used the identity $(\boldsymbol{q}-\boldsymbol{p})^\top\boldsymbol{W}_l^\top\boldsymbol{\delta}_l = \langle\boldsymbol{W}_l(\boldsymbol{q}-\boldsymbol{p}),\,\boldsymbol{\delta}_l\rangle$.

Applying the H\"older inequality in the form $|\langle \boldsymbol{u}, \boldsymbol{v} \rangle| \le \|\boldsymbol{u}\|_1 \|\boldsymbol{v}\|_\infty$ with $\boldsymbol{u} = \boldsymbol{W}_{l}(\boldsymbol{q} - \boldsymbol{p}) \in \mathbb{R}^{d_m}$ and $\boldsymbol{v} = \boldsymbol{\delta}_l \in \mathbb{R}^{d_m}$ yields the following bound on the first-order term, which together with \eqref{eq:kl_remainder} gives \eqref{eq:kl_bound}:
\begin{equation}
    \bigl| \langle \boldsymbol{W}_{l}(\boldsymbol{q} - \boldsymbol{p}),\, \boldsymbol{\delta}_l \rangle \bigr|
    \le \|\boldsymbol{W}_{l}(\boldsymbol{q} - \boldsymbol{p})\|_1 \cdot \|\boldsymbol{\delta}_l\|_\infty.
\end{equation}
Similarly, expanding the first-order inner product and applying the triangle inequality gives:
\begin{equation}
    \bigl| \langle \boldsymbol{W}_{l}(\boldsymbol{q} - \boldsymbol{p}),\, \boldsymbol{\delta}_l \rangle \bigr|
    = \Bigl| \sum_{i=1}^{d_m} \bigl[ \boldsymbol{W}_{l} (\boldsymbol{q} - \boldsymbol{p}) \bigr]_i \cdot (\boldsymbol{\delta}_l)_i \Bigr|
    \le \sum_{i=1}^{d_m} \bigl| \bigl[ \boldsymbol{W}_{l} (\boldsymbol{q} - \boldsymbol{p}) \bigr]_i \bigr| \cdot \bigl| (\boldsymbol{\delta}_l)_i \bigr|.
\end{equation}

Adding the remainder bound \eqref{eq:kl_remainder} proves \eqref{eq:element_bound}. Both bounds separate the unknown target mismatch $\boldsymbol{q} - \boldsymbol{p}$ from the observable feature perturbation $\boldsymbol{\delta}_l$ in the first-order term. Since $\boldsymbol{p}$ and $\boldsymbol{q}$ are probability vectors, we have $\|\boldsymbol{q} - \boldsymbol{p}\|_1 \le 2$, and the gradient-related factor $\|\boldsymbol{W}_{l}(\boldsymbol{q} - \boldsymbol{p})\|_1$ is bounded by $2\|\boldsymbol{W}_l\|_{1\to1}$, where $\|\boldsymbol{W}_l\|_{1\to1}=\max_j\sum_i|(\boldsymbol{W}_l)_{ij}|$. Together with the quadratic remainder, this gives a worst-case bound on the fixed-target KL loss change under the linear mapping. For token $j$, we apply the analysis with $\boldsymbol{\delta}_l=\Delta\boldsymbol{h}_{:,j}$, whose coordinate magnitudes are $|\Delta\boldsymbol{h}_{i,j}|$. The quadratic remainder quantifies the approximation error, while the fusion rule uses these observed feature magnitudes as its practical selection criterion. This first-order sensitivity analysis motivates our perturbation-driven fusion strategy in \cref{sec:module2}: by constructing the fusion mask $\boldsymbol{\mathcal{M}}_{\text{fusion}}$ from the magnitude $|\Delta\boldsymbol{h}_{i,j}|$ of the output discrepancy, we use coordinate-wise feature modification as a computationally feasible criterion for selective integration.
\end{proof}

\section{Experimental Settings}
\label{app:exp_settings}
In our experiments, ROSE is attached to the 26-th layer. Optimization is performed using the AdamW optimizer with a learning rate of $5 \times 10^{-4}$. Regarding the stopping criteria for an individual edit, the optimization loop continues until the loss drops below a threshold of $0.01$ or a maximum of 25 steps is reached. The threshold $k$ in $\boldsymbol{\mathcal{M}}_{\text{risk}}$ is set to $0.5$, the fusion parameter $\sigma=2$, and the expansion threshold $\epsilon=1 \times 10^{-5}$. 

\section{Impact of the Reference Anchor for Risk Identification}
\label{app:more_exp}

\begin{figure}[ht]
    \centering
    \includegraphics[width=0.9\linewidth]{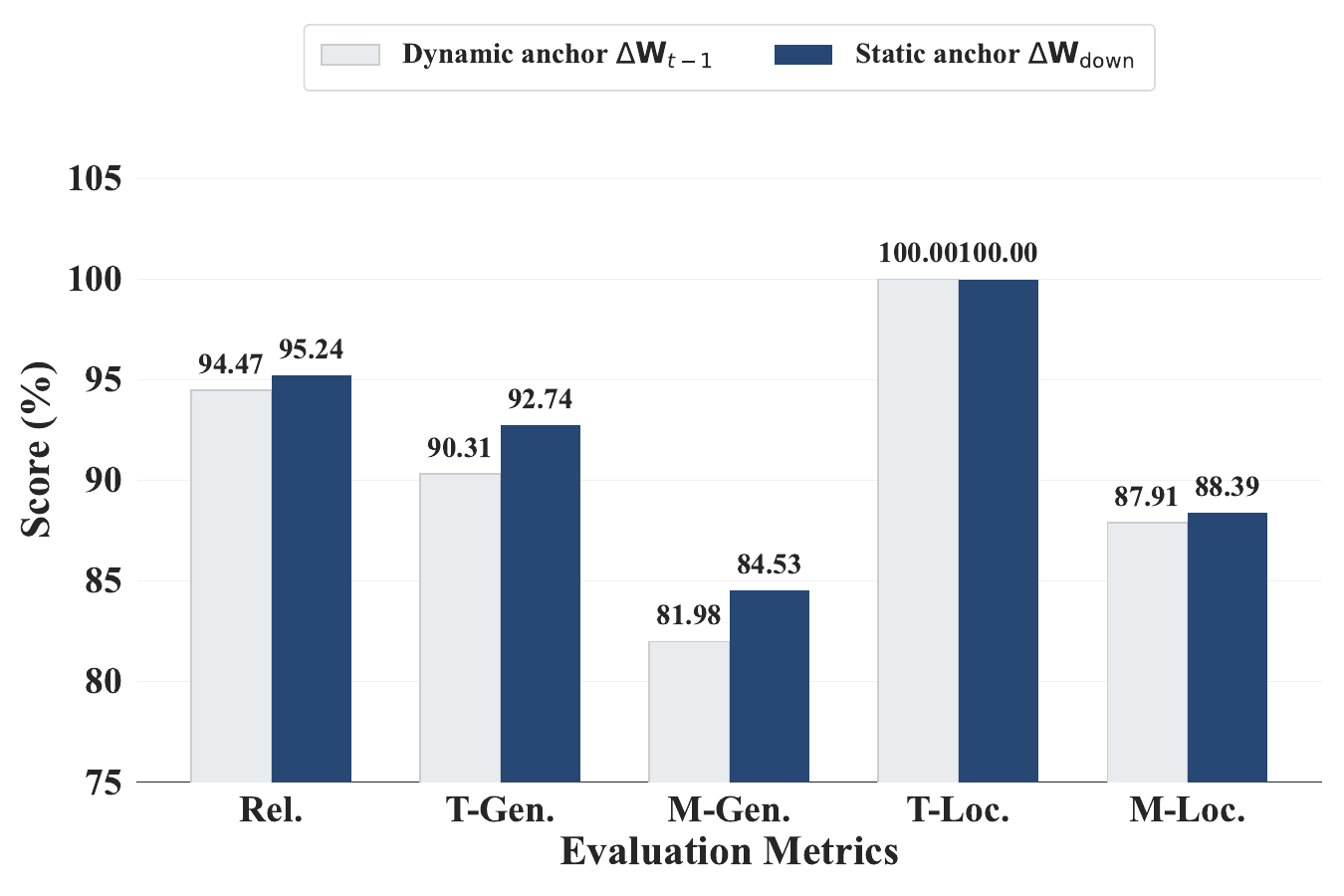} 
    \caption{Impact of the Reference Anchor for Risk Identification.}
    \label{fig:anchor}
\end{figure}

We examine how to choose the reference weight used to compute the gradient perturbation $\Delta \boldsymbol{W}$. In particular, we compare two strategies: the sequential anchor $\boldsymbol{W}_{t-1}$ and the \textit{static anchor} $\boldsymbol{W}_{\text{down}}$ derived from the backbone. As illustrated in \cref{fig:anchor}, employing the sequential anchor results in lower multimodal Locality. We ascribe this behavior to cumulative errors: as the model undergoes continuous updates, the reference point $\boldsymbol{W}_{t-1}$ gradually drifts away from the original pre-trained knowledge space. As a result, computing sensitivity relative to this shifted base fails to identify the subspaces that are essential for maintaining the model's inherent capabilities. In contrast, using the static anchor $\boldsymbol{W}_{\text{down}}$ offers a stable baseline. This helps the risk mask $\mathcal{M}_{\text{risk}}$ reliably safeguard the original knowledge structure throughout sequential editing, thereby supporting our design choice.

\section{Impact of Different MLLM Backbones}
\label{sec:backbone_scale}

To ensure the broad applicability and generalization of our approach, we extend the evaluation of ROSE beyond the default LLaVA-v1.5-7B model to encompass both different model architectures and a larger model scale. Specifically, we conduct comprehensive lifelong editing experiments on the MiniGPT-4 (7B) architecture, the Qwen2-VL (7B) model, and the larger LLaVA-v1.5-13B backbone.

First, we evaluate the performance of ROSE on the MiniGPT-4 (7B) model across the VLKEB, E-IC, and E-VQA benchmarks, as detailed in \cref{tab:minigpt4}. The experimental results clearly demonstrate the robust adaptability of our framework to different network structures. Even after a long stream of sequential edits, ROSE maintains high reliability and exceptional locality across all three benchmarks. This confirms that the risk-aware orthogonal projection and multi-granularity knowledge integration mechanisms can effectively mitigate catastrophic forgetting and preserve the model's fundamental visual-linguistic capabilities, regardless of the specific architectural nuances of the MLLM.

\begin{figure}[ht]
\centering
\includegraphics[width=0.9\textwidth]{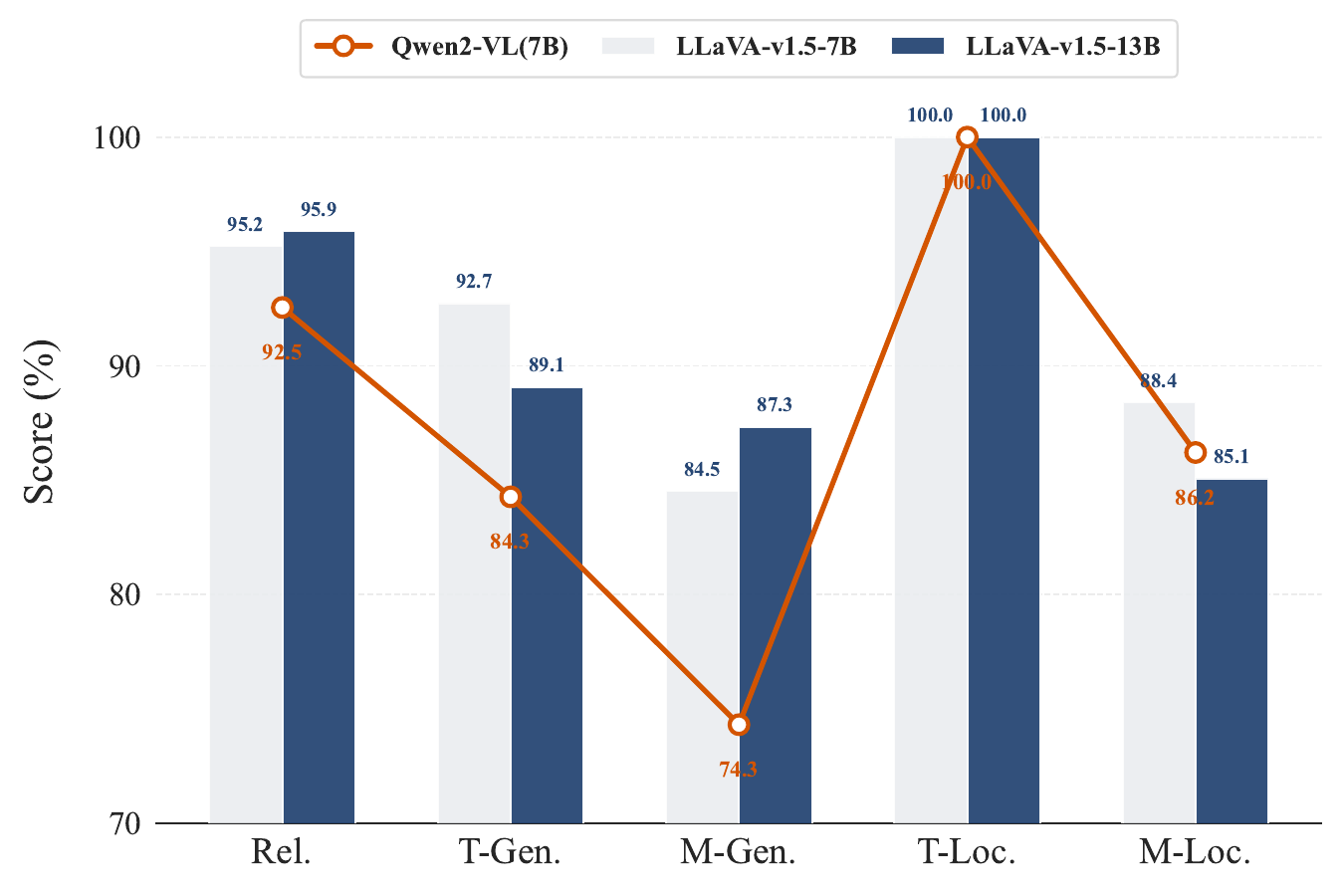}
\caption{Performance comparison across LLaVA-v1.5-7B, LLaVA-v1.5-13B, and Qwen2-VL-7B backbones after 1,000 sequential edits on the VLKEB benchmark.}
\label{fig:backbone_scale}
\end{figure}

Furthermore, to verify whether our mechanisms can effectively safeguard a significantly expanded parameter space and generalize to different model families, we scale our evaluation to the LLaVA-v1.5-13B backbone as well as the Qwen2-VL (7B) model. \cref{fig:backbone_scale} presents the performance comparison across these variants on the VLKEB benchmark after continuous edits. Notably, ROSE scales seamlessly to the larger 13B LLaVA model, yielding competitive results in the larger-model evaluation. Moreover, the results on the Qwen2-VL (7B) architecture further validate the versatility of our method across diverse model families, as it achieves highly competitive scores across all evaluation metrics and demonstrates strong robustness.

Overall, these extended evaluations indicate that the ROSE framework successfully mitigates catastrophic forgetting and maintains excellent performance across the evaluated model scales and architectures.

\begin{table}[t]
\caption{Performance comparison (\%) on MiniGPT-4 (7B), with baseline results taken from \citet{chen2025lifelong}. Models are assessed on five primary aspects: Reliability (Rel.), Text Generality (T-Gen.), Multimodal Generality (M-Gen.), Text Locality (T-Loc.), and Multimodal Locality (M-Loc.) after 1, 100, and 1000 sequential edits.}
\label{tab:minigpt4}
\renewcommand{\arraystretch}{1.15}
    \resizebox{\textwidth}{!}{ 
\begin{tabular}{ccccccccccccccccc}
\toprule
\multirow{3}{*}{Edits}  & \multicolumn{1}{c}{\multirow{3}{*}{Methods}} & \multicolumn{5}{c}{VLKEB}                                                                                                            & \multicolumn{5}{c}{E-IC}                                                                                                             & \multicolumn{5}{c}{E-VQA}                                                                                                            \\ \cmidrule(lr){3-7} \cmidrule(lr){8-12} \cmidrule(lr){13-17} 
                        & \multicolumn{1}{c}{}                         & \multicolumn{1}{c}{\multirow{2}{*}{Reliability}} & \multicolumn{2}{c}{Generality}  & \multicolumn{2}{c}{Locality}   & \multicolumn{1}{c}{\multirow{2}{*}{Reliability}} & \multicolumn{2}{c}{Generality}  & \multicolumn{2}{c}{Locality}  & \multicolumn{1}{c}{\multirow{2}{*}{Reliability}} & \multicolumn{2}{c}{Generality}  & \multicolumn{2}{c}{Locality}  \\
                        & \multicolumn{1}{c}{}                         & \multicolumn{1}{c}{}                     & T-Gen.         & M-Gen.         & T-Loc.          & M-Loc.       & \multicolumn{1}{c}{}                     & T-Gen.         & M-Gen.         & T-Loc.       & M-Loc.         & \multicolumn{1}{c}{}                     & T-Gen.         & M-Gen.         & T-Loc.       & M-Loc.         \\ \hline
\multirow{10}{*}{T=1}   & FT-L                                         & 82.17                                    & 81.64          & 78.45          & 98.45           & 75.06        & 60.40                                    & 59.47          & 52.79          & 99.95        & 92.46          & 93.85                                    & 86.25          & 88.58          & 99.47        & 84.50          \\
                        & FT-M                                         & 93.51                                    & 91.28          & 68.30          & 100.00          & 81.78        & 71.60                                    & 67.34          & 49.55          & 100.00        & 91.69          & 91.42                                    & 85.82          & 49.09          & 100.00        & 86.72          \\
                        & MEND                                         & 90.28                                    & 89.18          & 89.55          & 93.63           & 86.69        & 90.28                                    & 83.91          & 79.05          & 95.92        & 93.97          & 93.47                                    & 91.32          & 91.22          & 81.51        & 74.97          \\
                        & SERAC                                        & 89.28                                    & 89.06          & 87.35          & 64.88           & 12.72        & 85.09                                    & 78.96          & 81.70          & 83.36        & 24.50          & 87.66                                    & 69.24          & 85.96          & 82.11        & 24.11          \\
                        & TP                                           & 49.17                                    & 51.88          & 48.92          & 90.63           & 79.79        & 52.19                                    & 51.49          & 49.81          & 83.33        & 72.04          & 52.45                                    & 48.39          & 51.56          & 93.93        & 83.61          \\
                        & LTE                                          & 92.92                                    & 85.16          & 86.25          & 88.29           & 81.96        & 91.30                                    & 84.62          & 83.55          & 94.96        & 91.88          & 93.54                                    & 77.59          & 84.07          & 94.03        & 88.14          \\
                        & RECIPE                                       & 93.20                                    & 92.40          & 90.71          & 90.29           & 84.35        & 81.25                                    & 74.44          & 76.96          & 97.18        & 92.17          & 89.23                                    & 70.39          & 87.94          & 95.49        & 90.90          \\
                        & LEMoE                                        & 92.26                                    & 89.28          & 88.27          & 93.87           & 67.49        & 90.32                                    & 86.30          & 85.75          & 96.66        & 69.17          & 92.59                                    & 90.49          & 89.16          & 97.01        & 84.33          \\
                        & LiveEdit                                     & 96.41                                    & 95.56          & \textbf{92.85} & 100.00          & \textbf{100.00} & 86.84                                    & 83.82          & 76.95          & 100.00        & \textbf{100.00} & 94.90                                    & 92.63          & \textbf{92.06} & 100.00        & \textbf{100.00}          \\ \cline{2-17} 
                        \addlinespace[1pt] 
                        & \gc ROSE                                     & \gc \textbf{100.00}                      & \gc \textbf{100.00} & \gc 87.38      & \gc \textbf{100.00} & \gc 97.81    & \gc \textbf{100.00}                      & \gc \textbf{94.16} & \gc \textbf{88.29} & \gc \textbf{100.00} & \gc 96.79      & \gc \textbf{100.00}                      & \gc \textbf{97.25} & \gc 87.67      & \gc \textbf{100.00} & \gc 99.27      \\     
                        \addlinespace[1pt]  
                        \hline
\multirow{10}{*}{T=100} & FT-L                                         & 69.98                                    & 69.05          & 68.09          & 88.74           & 63.07        & 62.35                                    & 57.49          & 54.13          & 96.15        & 46.86          & 64.81                                    & 58.11          & 54.23          & 93.62        & 44.98          \\
                        & FT-M                                         & 60.03                                    & 60.18          & 57.00          & 100.00          & 51.24        & 58.93                                    & 54.35          & 47.84          & 100.00        & 36.80          & 66.06                                    & 51.17          & 47.24          & 100.00        & 36.65          \\
                        & MEND                                         & 38.85                                    & 40.02          & 37.99          & 72.82           & 68.89        & 51.92                                    & 53.96          & 48.24          & 86.02        & 83.88          & 23.32                                    & 20.75          & 21.42          & 47.27        & 42.60          \\
                        & SERAC                                        & 69.38                                    & 58.31          & 67.57          & 51.71           & 13.72        & 47.41                                    & 46.17          & 45.22          & 49.37        & 13.52          & 87.39                                    & 66.53          & 81.70          & 65.13        & 14.04          \\
                        & TP                                           & 39.60                                    & 42.52          & 39.14          & 36.15           & 43.68        & 39.60                                    & 39.64          & 38.48          & 17.29        & 26.04          & 19.46                                    & 20.32          & 22.03          & 41.85        & 32.91          \\
                        & LTE                                          & 79.45                                    & 63.55          & 79.46          & 84.18           & 79.42        & 50.47                                    & 49.57          & 48.17          & 88.12        & 92.87          & 85.05                                    & 71.80          & 79.61          & 87.06        & 86.10          \\
                        & RECIPE                                       & 71.54                                    & 60.04          & 70.35          & 85.57           & 86.85        & 47.27                                    & 46.06          & 45.12          & 89.52        & 94.25          & 88.26                                    & 67.74          & 81.48          & 91.77        & 88.23          \\
                        & LEMoE                                        & 44.75                                    & 43.33          & 41.96          & 73.90           & 60.38        & 52.85                                    & 50.20          & 48.11          & 92.08        & 49.63          & 27.84                                    & 26.38          & 26.73          & 68.45        & 33.73          \\
                        & LiveEdit                                     & 94.06                                    & \textbf{94.62} & \textbf{89.42} & 100.00          & \textbf{100.00} & 84.38                                    & 83.53          & 71.85          & 100.00        & \textbf{100.00} & 93.23                                    & \textbf{91.27} & \textbf{88.34} & 100.00        & \textbf{99.61} \\ \cline{2-17} 
                        \addlinespace[1pt]
                        & \gc ROSE                                     & \gc \textbf{95.42}                       & \gc 94.20      & \gc 78.29      & \gc \textbf{100.00} & \gc 98.38    & \gc \textbf{91.41}                       & \gc \textbf{85.41} & \gc \textbf{81.30} & \gc \textbf{100.00} & \gc 99.14      & \gc \textbf{94.97}                       & \gc 88.52      & \gc 83.45      & \gc \textbf{100.00} & \gc 99.07      \\ 
                        \addlinespace[1pt]
                        \hline
\multirow{10}{*}{T=1000}& FT-L                                         & 61.85                                    & 60.82          & 60.41          & 76.17           & 60.34        & 58.88                                    & 53.98          & 52.98          & 93.31        & 46.49          & 58.97                                    & 46.34          & 50.77          & 72.04        & 41.92          \\
                        & FT-M                                         & 51.26                                    & 54.35          & 51.13          & 100.00          & 50.72        & 52.65                                    & 47.90          & 47.25          & 100.00        & 33.69          & 51.18                                    & 42.79          & 40.08          & 100.00        & 37.28          \\
                        & MEND                                         & 42.35                                    & 45.07          & 42.68          & 66.98           & 62.32        & 49.30                                    & 51.46          & 46.70          & 81.03        & 82.53          & 31.84                                    & 26.60          & 33.98          & 43.08        & 44.90          \\
                        & SERAC                                        & 57.02                                    & 51.66          & 56.80          & 48.48           & 13.24        & 47.03                                    & 45.79          & 44.82          & 42.12        & 15.31          & 84.50                                    & 60.36          & 81.83          & 63.26        & 13.40          \\
                        & TP                                           & 23.84                                    & 25.00          & 23.54          & 18.79           & 25.58        & 25.07                                    & 24.50          & 24.09          & 14.77        & 18.05          & 9.24                                     & 8.96           & 10.25          & 20.54        & 17.03          \\
                        & LTE                                          & 64.28                                    & 56.91          & 64.07          & 83.96           & 78.59        & 51.31                                    & 49.25          & 49.19          & 88.19        & 90.53          & 81.93                                    & 66.81          & 75.96          & 88.24        & 78.95          \\
                        & RECIPE                                       & 58.18                                    & 52.31          & 58.14          & 85.49           & 83.84        & 47.02                                    & 45.76          & 44.85          & 89.80        & 92.26          & 85.33                                    & 61.79          & 76.35          & 90.33        & 89.08          \\
                        & LEMoE                                        & 39.48                                    & 38.24          & 36.52          & 82.35           & 58.01        & 50.61                                    & 47.35          & 47.19          & 95.11        & 47.62          & 25.07                                    & 23.87          & 26.68          & 89.11        & 50.50          \\
                        & LiveEdit                                     & \textbf{93.41}                           & 89.87          & \textbf{85.21} & 100.00          & \textbf{100.00} & 81.43                                    & 80.34          & 62.25          & 100.00        & \textbf{99.73} & \textbf{92.75}                           & \textbf{89.81} & \textbf{85.12} & 100.00        & \textbf{96.82} \\ \cline{2-17} 
                        \addlinespace[1pt]
                        & \gc ROSE                                     & \gc 91.12                                & \gc \textbf{89.92} & \gc 74.75      & \gc \textbf{100.00} & \gc 86.84    & \gc \textbf{86.90}                       & \gc \textbf{80.76} & \gc \textbf{70.93} & \gc \textbf{100.00} & \gc 97.01      & \gc 92.35                                & \gc 78.60      & \gc 77.63      & \gc \textbf{100.00} & \gc 91.23      \\ 
                        \addlinespace[1pt]
                        \hline
\end{tabular}
}
\end{table}

\section{Theoretical Discussions on Capacity Limits}
\label{sec:theoretical_discussions}

A fundamental theoretical boundary exists in projection-based lifelong learning methods. When the accumulated historical feature subspace reaches full rank $d_n$, maintaining absolute orthogonality to all historical directions permits only the zero update. Under such conditions, acquiring novel facts through further weight updates requires a trade-off with protecting all historical feature directions. To address this capacity limit, the ROSE framework can be naturally extended with a Singular Value Decomposition (SVD)-based compression of the historical feature matrix, whose columns contain historical feature vectors, retaining its leading left singular vectors as a reduced orthogonal basis.

According to the properties of SVD, discarding components associated with the smallest singular values yields an optimal low-rank approximation of the historical feature matrix in Frobenius or spectral norm. This form of controlled feature compression represents a mathematically sound compromise under strict finite-parameter constraints.

Our reported experiments evaluate sequences of up to $1,000$ sequential edits without capacity compression. Within this regime, the available update directions depend on the actual rank of the accumulated historical feature subspace. However, in extreme real-world deployments where continuous edits may scale to tens of thousands, this subspace may approach its dimensionality boundary. In such limiting scenarios, the aforementioned SVD-based truncation offers a possible extension for managing the finite subspace budget.

\section{Broader Impacts}
\label{sec:broader_impacts}

Our research on lifelong knowledge editing for Multimodal Large Language Models (MLLMs) presents significant positive societal impacts. By enabling models to dynamically update their internal knowledge without catastrophic forgetting, ROSE effectively mitigates the issue of hallucinations and outdated factual responses. This can help ensure that deployed AI systems can remain factual, reliable, and aligned with the evolving real world, which is crucial for safety-critical applications such as healthcare, legal consultation, and education.

However, we acknowledge the potential negative societal impacts and dual-use risks associated with this technology. The ability to precisely and efficiently modify an MLLM's knowledge could be exploited by malicious actors to intentionally inject disinformation, bias, or harmful content into open-source models. Since ROSE is designed to seamlessly integrate edits while preserving the model's general capabilities, such poisoned models might behave normally on standard queries while outputting manipulative content for specific triggers. To mitigate these risks, we recommend that practitioners implement robust fact-checking verification pipelines before applying any edits. Furthermore, maintaining a transparent and auditable log of the Knowledge Bank is essential for tracking and reverting any unauthorized or malicious modifications in public-facing AI systems.

\section{Algorithmic Implementation Details of ROSE}
\label{sec:algorithm_details}

To provide a clear and step-by-step overview of the ROSE framework, we summarize its complete execution procedure in \cref{alg:rose}. The pseudocode concisely outlines our two primary stages: the lifelong editing phase for safely absorbing new sequential edits, and the inference phase for accurate, perturbation-guided knowledge routing.

\clearpage

\begin{algorithm}[H]
\small
\setlength{\baselineskip}{10pt}
\caption{ROSE: Risk-Aware Orthogonal Subspace Navigation for Lifelong Editing}
\label{alg:rose}
\begin{algorithmic}[1]
\Require 
    Pre-trained MLLM backbone with down-projection matrix $\boldsymbol{W}_{\text{down}}$; 
    Sequential edit stream $\mathcal{D}_{\text{edit}} = \{e_1, e_2, \ldots, e_T\}$; 
    Hyperparameters $k, \epsilon, \tau_{\text{min}}, \tau, \tau_{\text{gap}}, \sigma$; Learning rate $\eta$.
\Ensure 
    Updated auxiliary weights $\boldsymbol{W}_t$; Updated basis $\boldsymbol{U}^{(t)}$; Knowledge Bank $\boldsymbol{\mathcal{K}}$.

\State \textbf{Initialize:} $\boldsymbol{W}_0 \leftarrow \boldsymbol{W}_{\text{down}}$; $\boldsymbol{U}^{(0)}_v, \boldsymbol{U}^{(0)}_p \leftarrow \emptyset$; $\boldsymbol{Q}^{(0)} \leftarrow \emptyset$; $\boldsymbol{\mathcal{K}}_v, \boldsymbol{\mathcal{K}}_p \leftarrow \emptyset$.

\vspace{0.1cm}
\Statex \textcolor{blue}{\textit{\#=== Phase 1: Lifelong Editing Phase (Risk-Guided Orthogonal Editing) ===\#}}
\For{each edit $e_t = (v_t, p_t, o_t) \in \mathcal{D}_{\text{edit}}$}
    
    \State Initialize current weights $\boldsymbol{W}_t \leftarrow \boldsymbol{W}_{t-1}$
    \While{not converged for $e_t$}
        \State $\Delta \boldsymbol{W} \leftarrow \boldsymbol{W}_t - \boldsymbol{W}_{\text{down}}$
        \State \textbf{Determine} the $(1-k)$-th quantile threshold $\psi_k$ of elements in $\{|\Delta \boldsymbol{W}_{i,j}|\}$
        \State Compute risk mask $\boldsymbol{\mathcal{M}}_{\text{risk}} \leftarrow \bigl[ \mathbb{I} \bigl( |\Delta \boldsymbol{W}_{i,j}| > \psi_k \bigr) \bigr]_{i,j}$ \Comment{Eq. 5}
        \State Compute loss $\mathcal{L}$ for $e_t$
        \State Calculate raw gradient $\nabla_{\boldsymbol{W}_t} \mathcal{L}$
        \If{$t = 1$}
            \State Use the raw gradient for the update
        \ElsIf{$\boldsymbol{Q}^{(t-1)}$ is empty}
            \State Project gradient: $\nabla_{\boldsymbol{W}_t} \mathcal{L} \leftarrow \nabla_{\boldsymbol{W}_t} \mathcal{L} \odot \boldsymbol{\mathcal{M}}_{\text{risk}}$
        \Else
            \State Project gradient: $\nabla_{\boldsymbol{W}_t} \mathcal{L} \leftarrow \bigl( \nabla_{\boldsymbol{W}_t} \mathcal{L} - \nabla_{\boldsymbol{W}_t} \mathcal{L} \boldsymbol{Q}^{(t-1)} (\boldsymbol{Q}^{(t-1)})^\top \bigr) \odot \boldsymbol{\mathcal{M}}_{\text{risk}}$ \Comment{Eq. 6}
        \EndIf
        \State Update weights: $\boldsymbol{W}_t \leftarrow \boldsymbol{W}_t - \eta \nabla_{\boldsymbol{W}_t} \mathcal{L}$
    \EndWhile
    
    \State Extract decoupling features $\boldsymbol{x}'_v, \boldsymbol{x}'_p$ from $v_t, p_t$
    \For{modality $*\in \{v, p\}$}
        \State Normalize $\hat{\boldsymbol{x}}_* \leftarrow \boldsymbol{x}'_*/\|\boldsymbol{x}'_*\|_2$
        \If{$\boldsymbol{U}_*^{(t-1)}$ is empty}
            \State $\tilde{\boldsymbol{x}}_* \leftarrow \hat{\boldsymbol{x}}_*$
        \Else
            \State $\tilde{\boldsymbol{x}}_* \leftarrow \hat{\boldsymbol{x}}_* - \boldsymbol{U_*}^{(t-1)} (\boldsymbol{U_*}^{(t-1)})^\top \hat{\boldsymbol{x}}_*$ \Comment{Eq. 2}
        \EndIf
        \If{$\|\tilde{\boldsymbol{x}}_*\|_2 > \epsilon$}
            \State $\boldsymbol{U}_{*}^{(t)} \leftarrow[\boldsymbol{U}_{*}^{(t-1)}, \tilde{\boldsymbol{x}}_* / \|\tilde{\boldsymbol{x}}_*\|_2]$ \Comment{Eq. 3}
        \Else
            \State $\boldsymbol{U}_{*}^{(t)} \leftarrow \boldsymbol{U}_{*}^{(t-1)}$
        \EndIf
        \State Compute semantic prototype $\boldsymbol{k}_*$ and update Bank $\boldsymbol{\mathcal{K}}_* \leftarrow [\boldsymbol{\mathcal{K}}_*, \boldsymbol{k}_*]$ \Comment{Eq. 7}
    \EndFor
    
    \State Merge basis $\boldsymbol{U}_{\text{merged}}^{(t)} \leftarrow [\boldsymbol{U}_v^{(t)}, \boldsymbol{U}_p^{(t)}]$
    \State $\boldsymbol{Q}^{(t)}, \boldsymbol{R}^{(t)} \leftarrow \text{QR}(\boldsymbol{U}_{\text{merged}}^{(t)})$ \Comment{Eq. 4}
\EndFor

\vspace{0.1cm}
\Statex \textcolor{blue}{\textit{\#=== Phase 2: Inference Phase (Multi-Granularity Knowledge Integration) ===\#}}
\Procedure{Inference}{$\boldsymbol{x}$}
    \State Extract global query vectors $\boldsymbol{q}_v, \boldsymbol{q}_p$ from input $\boldsymbol{x}$
    \State Compute alignment scores $s_v \leftarrow \max_j \langle \boldsymbol{q}_v, \boldsymbol{k}_{v,j}\rangle$, \ $s_p \leftarrow \max_j \langle \boldsymbol{q}_p, \boldsymbol{k}_{p,j}\rangle$ \Comment{Eq. 8}
    \State Determine gate  $\mathcal{G} \leftarrow \mathbb{I}( 
        \bigwedge\limits_{*\in\{v,p\}} (s_* > \tau_{\min}) 
        \land
        ( \bigvee\limits_{*\in\{v,p\}} (s_* > \tau)\lor |s_v - s_p| < \tau_{\text{gap}}
    ))$ \Comment{Eq. 9}
    
    \State Compute backbone output $\boldsymbol{h}_{\text{down}} \leftarrow \boldsymbol{W}_{\text{down}}\boldsymbol{x}$
    \If{$\mathcal{G} == 1$} \Comment{Route to Edited Knowledge Branch}
        \State Compute auxiliary output $\boldsymbol{h}_t \leftarrow \boldsymbol{W}_t\boldsymbol{x}$
        \State Calculate perturbation $\Delta \boldsymbol{h} \leftarrow \boldsymbol{h}_t - \boldsymbol{h}_{\text{down}}$
        \State Construct fusion mask $\boldsymbol{\mathcal{M}}_{\text{fusion}} \leftarrow \left[\mathbb{I}(|\Delta\boldsymbol{h}_{i,j}| > \mathbb{E}[|\Delta\boldsymbol{h}|] + \sigma \cdot \text{Std}[|\Delta\boldsymbol{h}|])\right]_{i,j}$ \Comment{Eq. 10}
        \State $\boldsymbol{\mathcal{M}}_{\text{gate}} \leftarrow \mathcal{G} \cdot \boldsymbol{\mathcal{M}}_{\text{fusion}}$
        \State $\boldsymbol{h}_{\text{final}} \leftarrow (1 - \boldsymbol{\mathcal{M}}_{\text{gate}}) \odot \boldsymbol{h}_{\text{down}} + \boldsymbol{\mathcal{M}}_{\text{gate}} \odot \boldsymbol{h}_t$ \Comment{Eq. 11}
    \Else \Comment{Route to Pre-trained Backbone}
        \State $\boldsymbol{h}_{\text{final}} \leftarrow \boldsymbol{h}_{\text{down}}$
    \EndIf
    \State \Return $\boldsymbol{h}_{\text{final}}$
\EndProcedure
\end{algorithmic}
\end{algorithm}

\end{document}